\documentclass[10pt,twocolumn,letterpaper]{article}

\usepackage[pagenumbers]{cvpr} % To force page numbers, e.g. for an arXiv version

\usepackage{graphicx}
\usepackage{amsmath}
\usepackage{amssymb}
\usepackage{booktabs}
\usepackage{caption,subcaption}
\newcommand{\tabincell}[2]{\begin{tabular}{@{}#1@{}}#2\end{tabular}}
\usepackage{multirow}
\usepackage{overpic}
\usepackage{bbding}

\usepackage[pagebackref,breaklinks,colorlinks]{hyperref}

\def\eg{\emph{e.g}\onedot} 
\def\ie{\emph{i.e}\onedot}

\makeatother

\usepackage[capitalize]{cleveref}
\crefname{section}{Sec.}{Secs.}
\Crefname{section}{Section}{Sections}
\Crefname{table}{Table}{Tables}
\crefname{table}{Tab.}{Tabs.}
\Crefname{equation}{Equation}{Equations}
\crefname{equation}{Eqn.}{Eqns.}

\begin{document}

\title{Benchmark Dataset and Effective Inter-Frame Alignment for \\ Real-World Video Super-Resolution}

\author{Ruohao Wang$^1$, Xiaohui Liu$^1$, Zhilu Zhang$^1$, Xiaohe Wu$^1$, Chun-Mei Feng$^2$, Lei Zhang$^3$, Wangmeng Zuo$^1$$^{(}$\Envelope$^)$\\
$^1$Harbin Institute of Technology, China\\
$^2$Institute of High Performance Computing, A*STAR, Singapore\\ 
$^3$The Hong Kong Polytechnic University, China\\
{\tt\small rhwangHIT@outlook.com, lxh720199@gmail.com, cszlzhang@outlook.com, xhwu.cpsl.hit@gmail.com} 
\\
{\tt\small strawberry.feng0304@gmail.com, cslzhang@comp.polyu.edu.hk, wmzuo@hit.edu.cn}}
\maketitle

%%%%%%%%% ABSTRACT
\begin{abstract}
Video super-resolution (VSR) aiming to reconstruct a high-resolution (HR) video from its low-resolution (LR) counterpart has made tremendous progress in recent years. However, it remains challenging to deploy existing VSR methods to real-world data with complex degradations. On the one hand, there are few well-aligned real-world VSR datasets, especially with large super-resolution scale factors, which limits the development of real-world VSR tasks. On the other hand, alignment algorithms in existing VSR methods perform poorly for real-world videos, leading to unsatisfactory results. As an attempt to address the aforementioned issues, we build a real-world $\times$4 VSR dataset, namely MVSR4$\times$, where low- and high-resolution videos are captured with different focal length lenses of a smartphone, respectively. Moreover, we propose an effective alignment method for real-world VSR, namely EAVSR. EAVSR takes the proposed multi-layer adaptive spatial transform network (MultiAdaSTN) to refine the offsets provided by the pre-trained optical flow estimation network. Experimental results on RealVSR and MVSR4$\times$ datasets show the effectiveness and practicality of our method, and we achieve state-of-the-art performance in real-world VSR task. The dataset and code will be available at \url{https://github.com/HITRainer/EAVSR}.
\end{abstract}

%%%%%%%%% BODY TEXT
\section{Introduction}
\label{sec:intro}
\begin{figure}[t]
    \centering
        \centering
        \includegraphics[width=\linewidth]{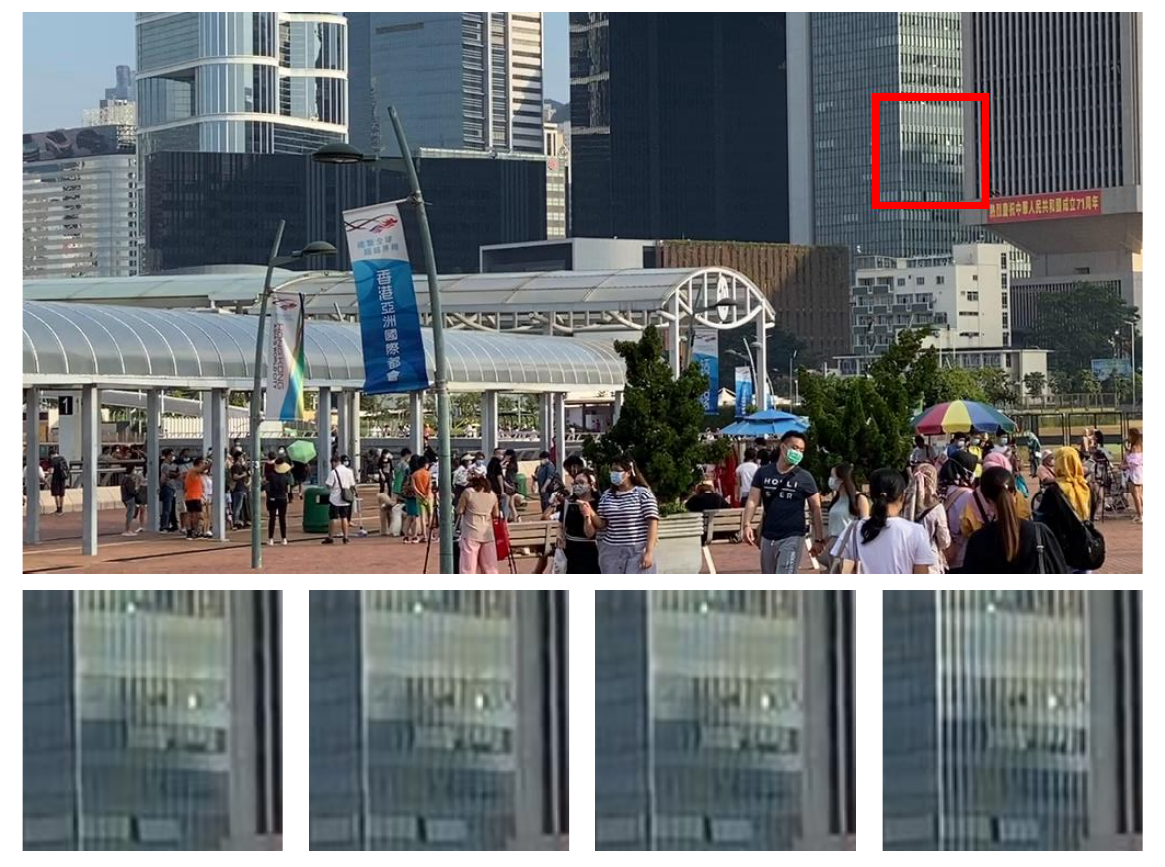}
    \put(-227,-8){\small ETDM~\cite{Isobe_2022_CVPR}}
    \put(-170,-8){\small TTVSR~\cite{Liu_2022_CVPR}}
    \put(-119,-8){\small  BasicVSR++~\cite{chan2021basicvsr++}}
    \put(-42,-8){Ours}
    \caption{Super-resolution results on a real-world video. In comparison with other VSR algorithms (\ie, ETDM~\cite{Isobe_2022_CVPR}, TTVSR~\cite{Liu_2022_CVPR} and BasicVSR++~\cite{chan2021basicvsr++}), our method restores more textures. Please zoom in for more details.}
    \label{fig:figure1} % is more photo-realistic.
    % \vspace{-2mm}
\end{figure}
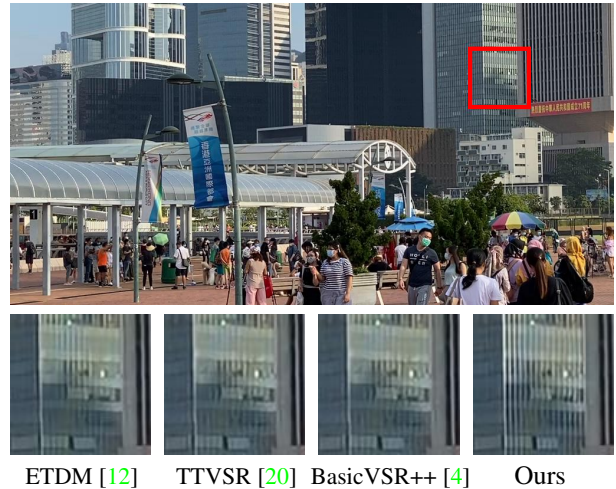

In comparison with single image super-resolution (SR) that reconstructs a high-resolution (HR) image from its low-resolution (LR) counterpart, video super-resolution (VSR) can further exploit valuable information from neighboring frames to recover richer textures and more details.
Thus, the inter-frame temporal alignment plays a significant role in VSR.
Driven by the development of convolutional neural networks (CNNs), especially the alignment methods~\cite{ranjan2017optical,dai2017deformable,zhu2019deformable}, tremendous progress has been made in VSR task. 

However, it is challenging for existing VSR methods to be effective in real-world applications.
On the one hand, most VSR methods are  trained only with synthetic ({\em e.g.}, bicubic downsampling) LR videos. 
Although RealBasicVSR~\cite{chan2022investigating}  adopts second-order order degradation model~\cite{wang2021real} and video compression to synthesize LR frames, its generalization capability on real-world LR videos is still limited as real degradations are more complex and changeable.
A natural solution is to capture real-world video pairs and take them to train the network.
Yang \textit{et al}.~\cite{yang2021real} has proposed a real-world VSR dataset with $\times{2}$ scale factor, but larger scale factor (\textit{e.g.}, $\times4$) datasets are wanting.
In addition, it is also more difficult to collect and process VSR datasets with larger scale-factor.

On the other hand, alignment algorithms in existing VSR methods perform poorly for real-world videos, leading to unsatisfactory results.
For example, TOF~\cite{xue2019video} and BasicVSR~\cite{chan2021basicvsr} estimate the offset between neighboring frames by optical flow network~\cite{ranjan2017optical} pre-trained on clean images. 
When facing low-quality real-world frames, the methods may not predict accurate offsets.
EDVR~\cite{wang2019edvr} and TDAN~\cite{tian2020temporally} introduce deformable convolution network (DCN)~\cite{dai2017deformable} for inter-frame alignment. 
Although such alignment methods are task-oriented and have more flexibility, it is easy to lead to the instability of network training~\cite{chan2020understanding}.
Chan {\em et al.}~\cite{chan2021basicvsr++} combine the optical flow network and DCN together, but it is still not effective enough for aligning real-world frames and can be further improved.

As an attempt to address the aforementioned issues, we build a new real-world video super-resolution dataset with  $\times4$ scale factor by mobile phone (\ie, Huawei P50 Pro), named MVSR4$\times$. 
The dual-view function in the smartphone is leveraged to take video frames with different focal length lenses with the same frequency almost synchronously. 
We collect the videos in different moving speeds, lightness, and scenes to guarantee data diversity.
In addition, it is inevitable that position misalignment and color inconsistency exist between images with different focal lengths.
To mitigate the adverse effect of these discrepancies on VSR model learning, we apply color correction and position alignment algorithms~\cite{shi2016real,sun2018pwc} not only in data pre-processing but also in the training phase. 

Furthermore, we propose an effective alignment method for real-world VSR, namely EAVSR, where a multi-layer adaptive spatial transform network (MultiAdaSTN) is proposed to refine inter-frame offsets. 
Specifically, MultiAdaSTN consists of ResflowNet and DeformNet.
Considering that the pre-trained optical flow network can provide a rough offset for neighboring frames, we utilize it to estimate a basic flow preliminarily. 
Noted that the basic flow is not completely accurate, and only taking DCN~\cite{dai2017deformable} to refine it (like BasicVSR++~\cite{chan2021basicvsr++}) may be inadequate as the optimization of DCN is hard.
Instead, we first compute a residual flow relative to the basic flow by ResflowNet, which introduces AdaSTN~\cite{zhang2022selfsupervised} and works in a pyramid coarse-to-fine manner. 
The combination of basic and residual flow can be considered as a more precise optical flow.
Then DeformNet is present to refine the optical flow to get ultimate inter-frame offsets.
Different from DCN, DeformNet utilizes the modified AdaSTN (\ie, AdaSTN v2) to predict the offsets and masks of deformable convolution v2~\cite{zhu2019deformable}.

The temporal propagation of our EAVSR adopts the bidirectional scheme of BasicVSR~\cite{chan2021basicvsr}. 
And we also introduce second-order grid propagation proposed by BasicVSR++~\cite{chan2021basicvsr++} into EAVSR to get EAVSR+.
Experiments are conducted on RealVSR~\cite{yang2021real} and MVSR4$\times$ datasets.
The results show the effectiveness and practicality of our proposed method, and it achieves state-of-the-art performance in real-world VSR task. 

To sum up, the main contributions of this work include:
\begin{itemize}
    \item We construct a benchmark dataset MVSR4$\times$, which compensates for the lack of paired real-world video datasets with large scale factors.
    \item We propose an effective alignment method EAVSR towards real-world data, where a multi-layer adaptive spatial transformer network (MultiAdaSTN) is designed elaborately to refine the inter-frame offsets.
    \item Experiments on RealVSR dataset and MVSR4$\times$ datasets show that our method achieves state-of-the-art performance in real-world VSR task.
\end{itemize}

\section{Related Work} \label{sec:relate}

\subsection{Degradation Model}

Most previous VSR methods~\cite{huang2017video,tao2017detail,sajjadi2018frame,xue2019video,li2019fast,haris2019recurrent,wang2020deep,isobe2020revisiting,isobe2020video,chan2020understanding,tian2020temporally,chan2021basicvsr,chan2021basicvsr++} focus on exploring inter-frame alignment and restoration networks based on synthetic datasets (\eg, REDS~\cite{nah2019ntire} and Vimeo-90K~\cite{xue2019video}), where LR videos are generated by bicubic down-sampling HR ones.
While significant performance has been gained on the synthetic datasets, the generalized results on real-world data are very limited due to the degradation gap.
For bridging this gap, RealBasicVSR~\cite{chan2022investigating} makes use of a second-order order degradation model~\cite{wang2021real} and video compression to synthesize LR frames, but its generalization capability on real-world LR videos is still unsatisfactory as real degradations are more complex and changeable.
Actually, to build VSR models for real scenarios, collecting real-world paired LR-HR videos is essential.
Recently, Yang \textit{et al}.~\cite{yang2021real} build a ${\times}$2 real-world video super-resolution dataset RealVSR by capturing paired LR-HR video sequences using the multi-camera system of iPhone 11 Pro Max.
In this work, to promote the development of real-world VSR, we construct a ${\times }$4 real-world video super-resolution dataset, namely MVSR$4\times$.

\subsection{Inter-Frame Alignment}
% \noindent{\textbf{Alignment.}}

Frame alignment is of great importance to use the complementary information across frames effectively in video restoration~\cite{chan2021basicvsr,pan2020cascaded,wronski2019handheld,xue2019video}.
Optical flow networks and their improved variants are extensively studied and deployed for motion estimation in VSR.
TOFlow~\cite{xue2019video} proposes a task-oriented flow module, which can be jointly trained with a video processing network.
OFRNet~\cite{wang2020deep} introduces an optical flow reconstruction network to infer HR optical flow in a coarse-to-fine manner for performing motion compensation.
However, optical flow networks can't deal with occlusions and sometimes make inaccurate predictions, impairing the restoration performance.

To alleviate the above issues, recent researchers aim to exploit deformable convolution network (DCN)~\cite{dai2017deformable} to align frames of videos.
For instance, TDAN~\cite{tian2020temporally} leverages DCN to perform alignment at the feature level instead of explicit motion estimation or image warping.
Inspired by TDAN, EDVR~\cite{wang2019edvr} deploys a pyramid cascading and deformable convolutions module to tackle the large motion.
Although DCN-based alignment is task-oriented with more flexibility, it shows poor training stability. 
Recently, BasicVSR++\cite{chan2021basicvsr++} proposes a flow-guided deformable alignment, but it is not effective enough for aligning neighboring frames from the real world.

%

% \textcolor{red}{transformer-based alignment}
Except for explicitly deploying the inter-frame alignment module, many algorithms use an indirect approach to fuse inter-frame information.
FSTRN~\cite{li2019fast} attempts to use 3D convolution networks to transfer temporal information.
ETDM~\cite{Isobe_2022_CVPR} computes the temporal difference between frames, divides those pixels into two subsets according to the level of difference, and uses different encoder networks to learn the distribution of these two differences separately.
In addition, with the development of Transformer, some methods achieve feature fusion by calculating the similarity of patches between frames.
TTVSR~\cite{Liu_2022_CVPR} leverages the self-attention module to calculate and focus on the relevant spatio-temporal trajectories.
MANA~\cite{MANA} designs a robust non-local attention mechanism that allows frames existing misalignment.
However, these methods are generally computationally expensive.

\section{Real-World $\times$4 VSR Dataset}
\label{sec:dataset}

\begin{figure}[t!]
    % \vspace{-3mm}
    \centering
        \centering
        \includegraphics[width=\linewidth]{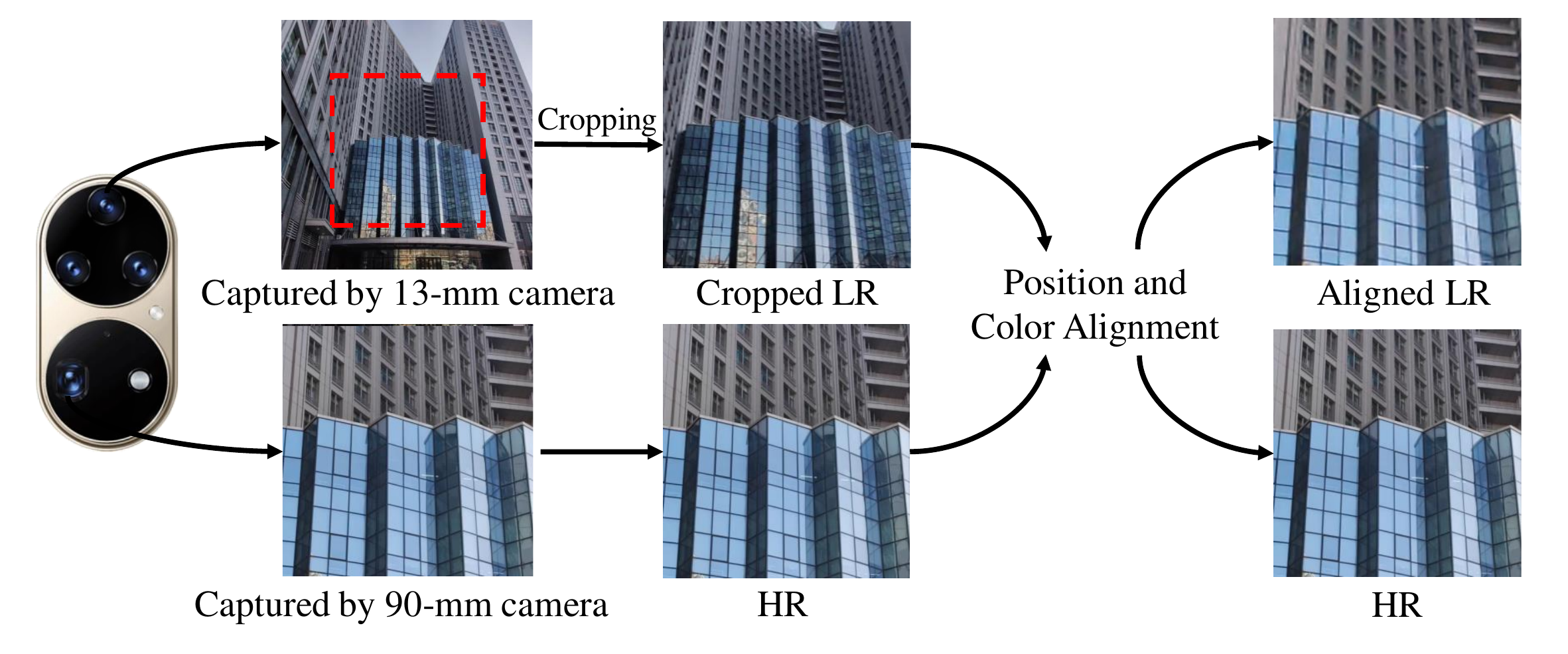}
    % \vspace{-2.5mm}
    \caption{Pre-processing pipeline of our MVSR4$\times$ dataset.}
    \label{fig:process} % is more photo-realistic.
\end{figure}
\begin{table}[t]
  \small
  \caption{Comparison of existing VSR datasets.}
  \vspace{-4mm}
  % \centering\noindent
  \centering%
  \begin{center}
    \begin{tabular}{lcccc}
      \toprule
      Dataset
      & \tabincell{c}{Real-\\World}
      & \tabincell{c}{Ground-\\Truth}
      & Aligned
      & \tabincell{c}{Scale\\Factor}
      \\
        \midrule
        Vimeo-90K~\cite{xue2019video}   & $\times$ & \checkmark & \checkmark & 4\\
        REDS~\cite{nah2019ntire}        & $\times$ & \checkmark & \checkmark & 4\\
        VideoLQ~\cite{chan2022investigating}        & \checkmark & $\times$ & - & -\\
        RealMCVSR~\cite{Lee_2022_CVPR}        & \checkmark & $\times$ & $\times$ & 4\\
        RealVSR~\cite{yang2021real}        & \checkmark & \checkmark & \checkmark & 2\\
        MVSR4$\times$        & \checkmark & \checkmark & \checkmark & 4 \\
        % \hline
      \bottomrule
    \end{tabular}
    \end{center}
  \label{tab:dataset}
  \vspace{-6mm}
\end{table}

\begin{figure*}[t!]
    \centering
    \small
    % \vspace{-6mm}
        \centering
        \subcaptionbox*{}
        {
            \includegraphics[width=.107\linewidth]{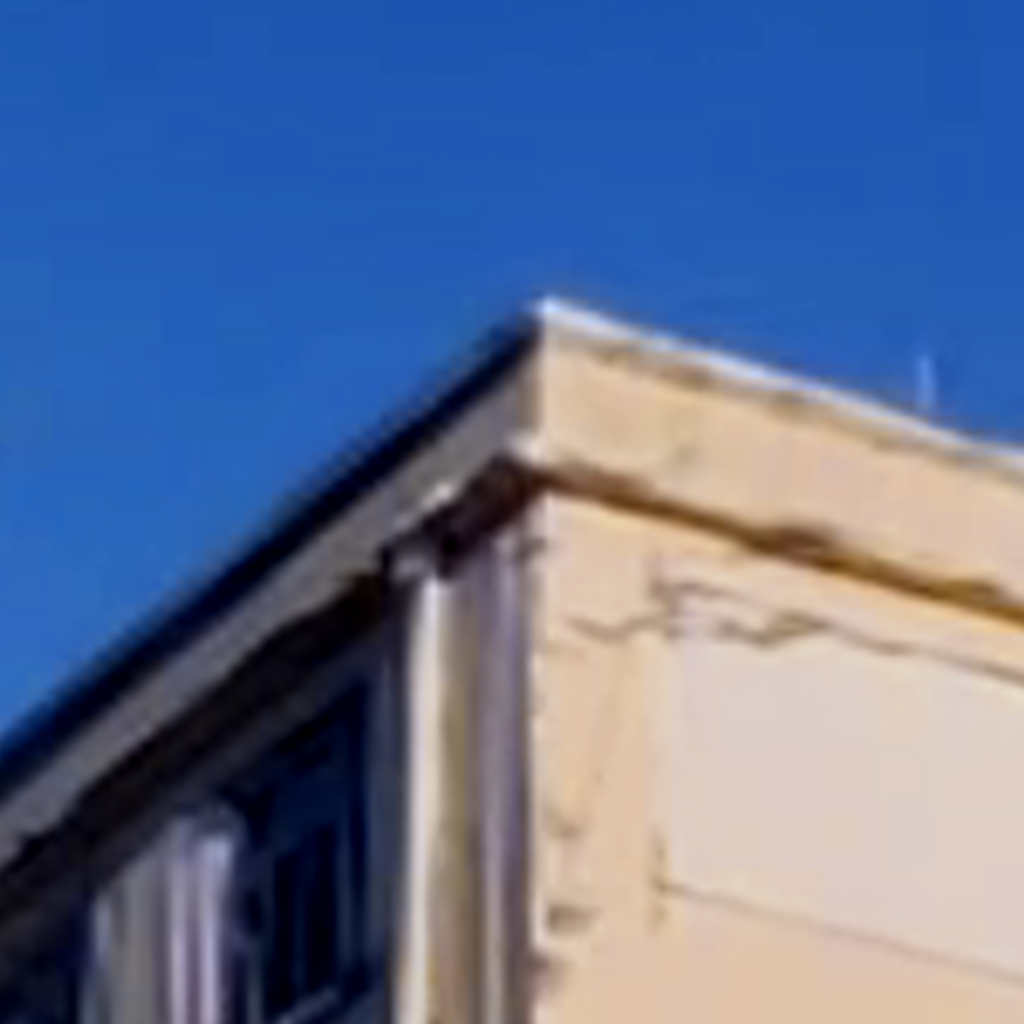}
        \hspace{-1mm}
            \includegraphics[width=.107\linewidth]{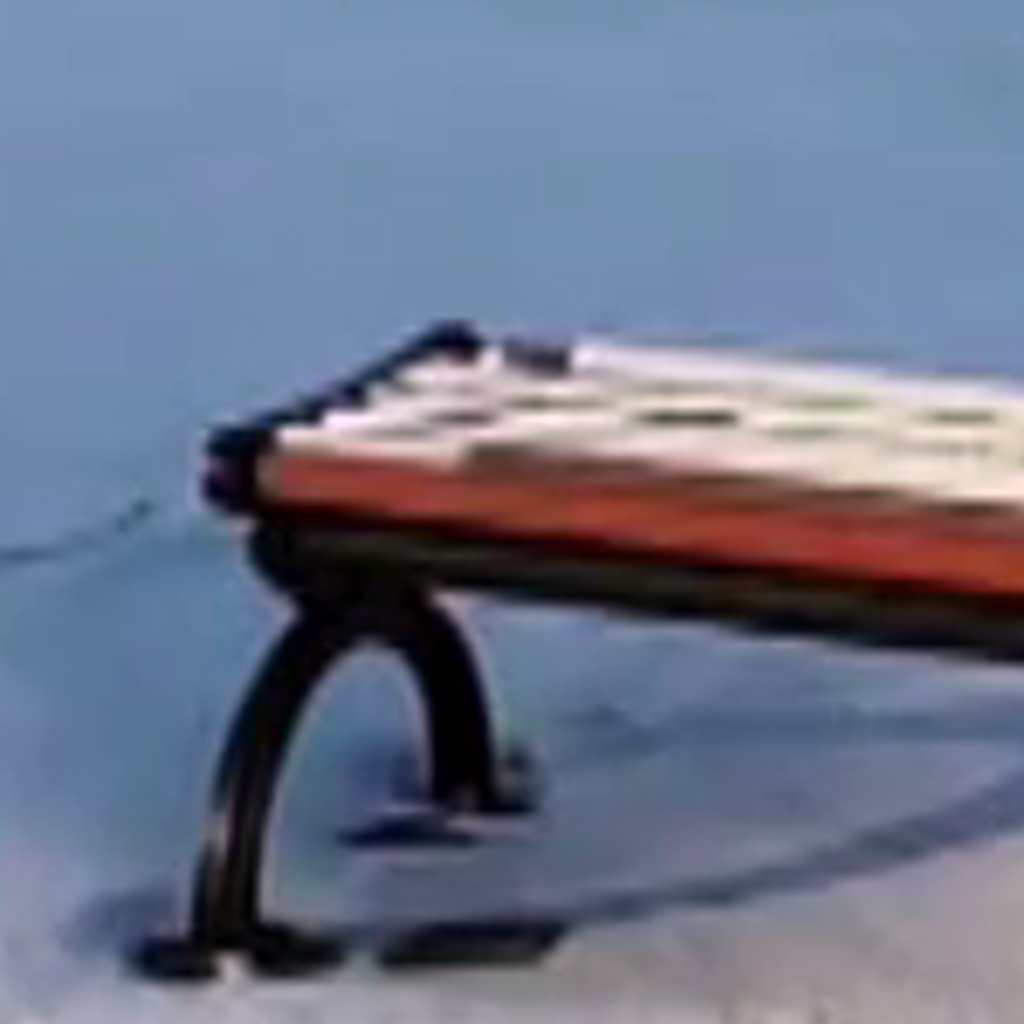}
        \hspace{-1mm}
            \includegraphics[width=.107\linewidth]{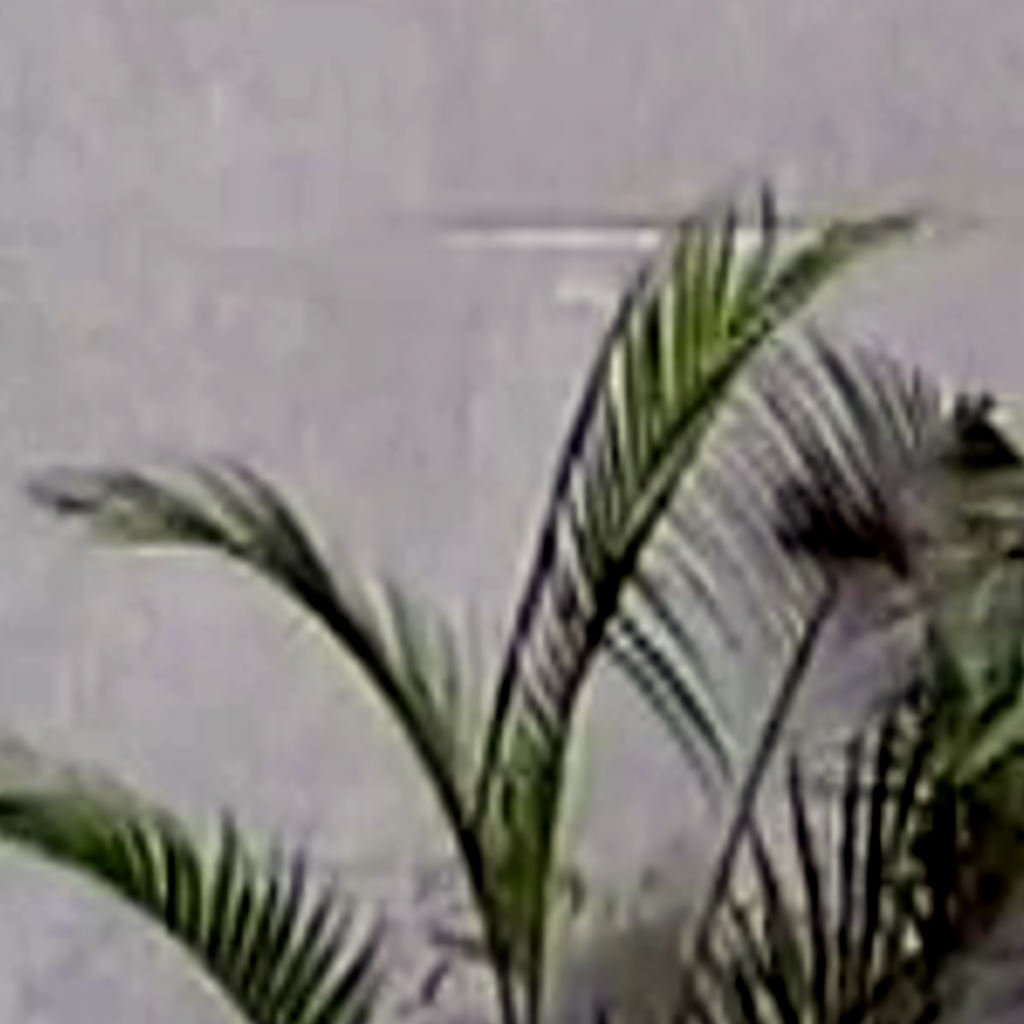}
        \hspace{-1mm}
            \includegraphics[width=.107\linewidth]{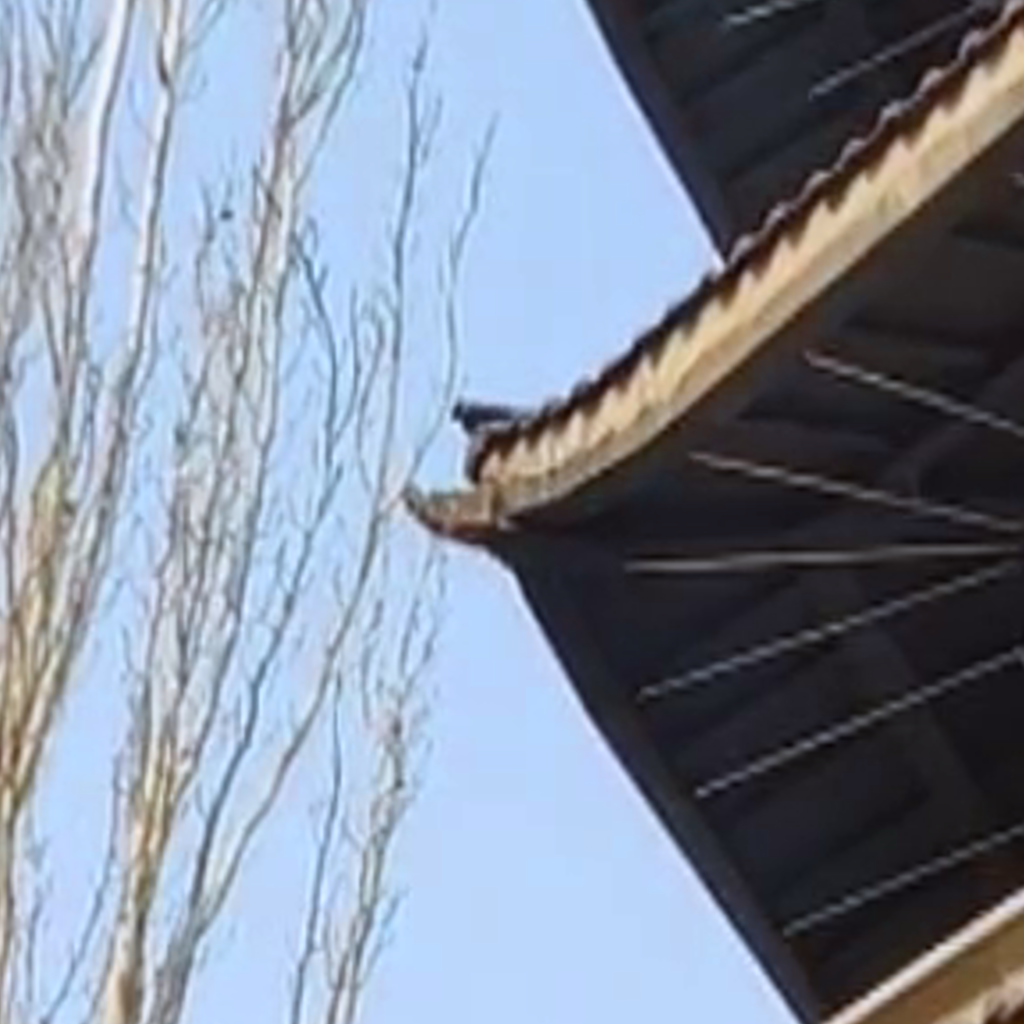}
        \hspace{-1mm}
            \includegraphics[width=.107\linewidth]{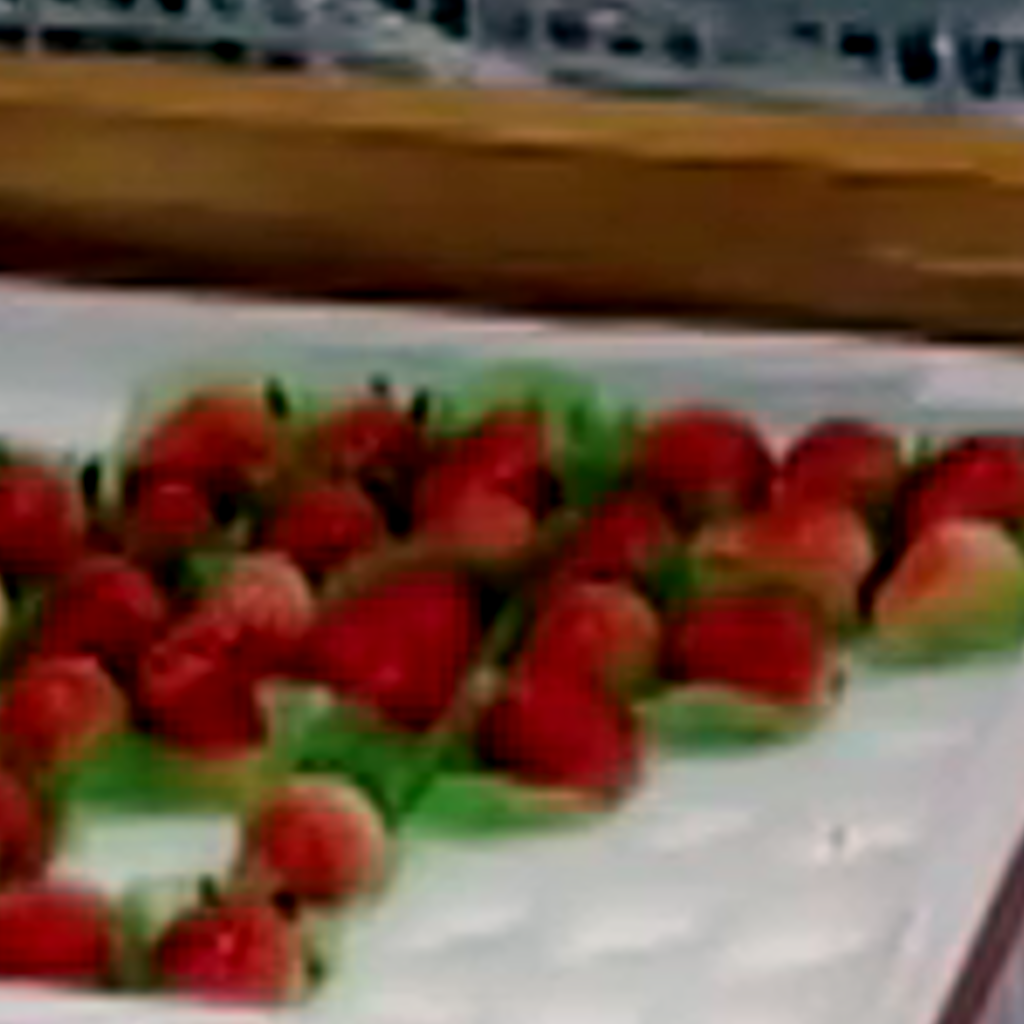}
        \hspace{-1mm}
            \includegraphics[width=.107\linewidth]{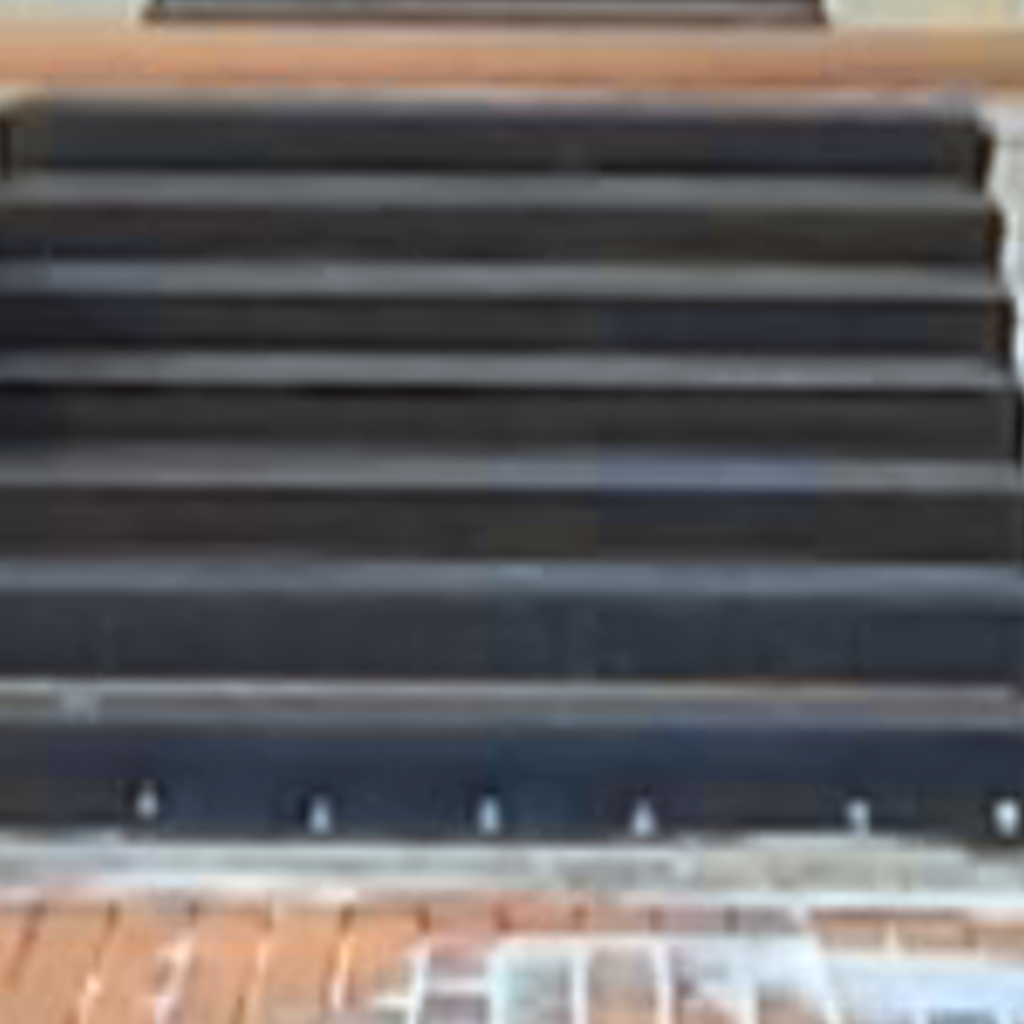}
        \hspace{-1mm}
            \includegraphics[width=.107\linewidth]{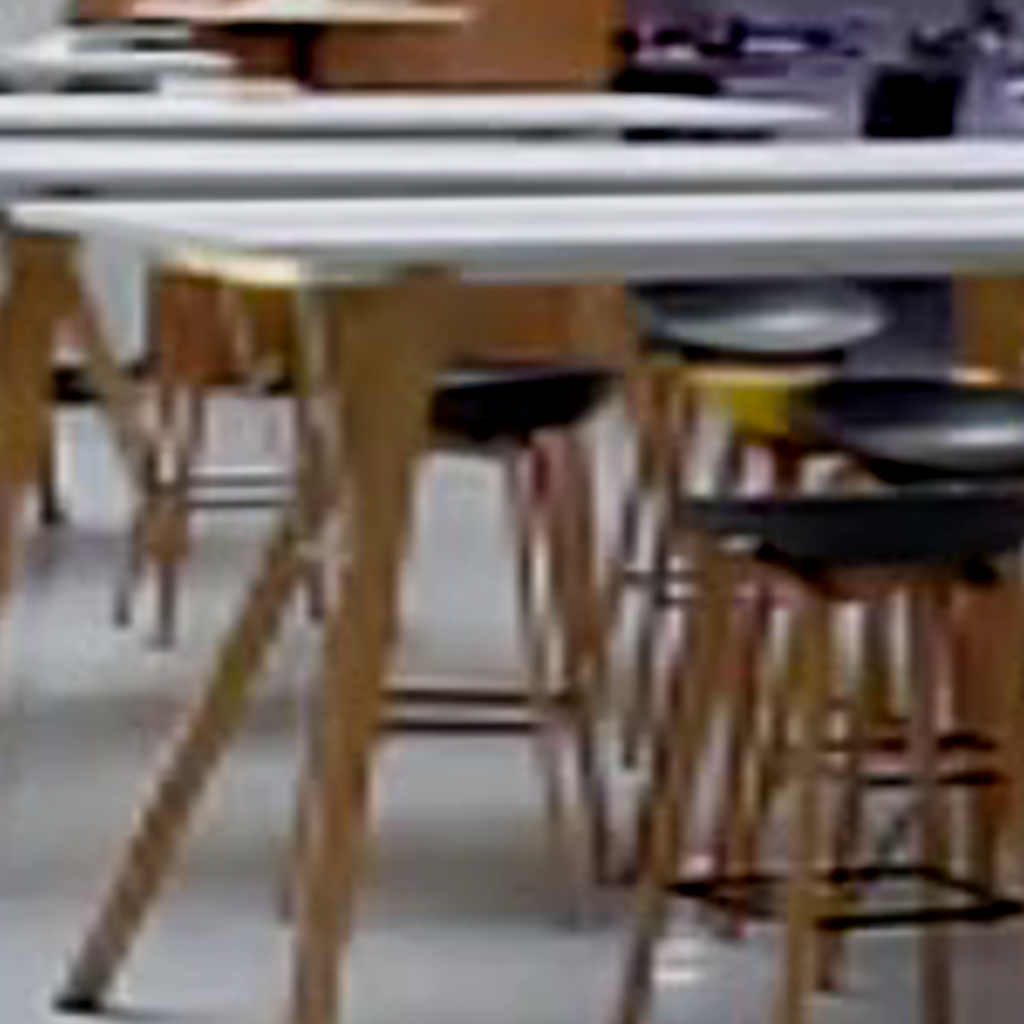}
        \hspace{-1mm}
            \includegraphics[width=.107\linewidth]{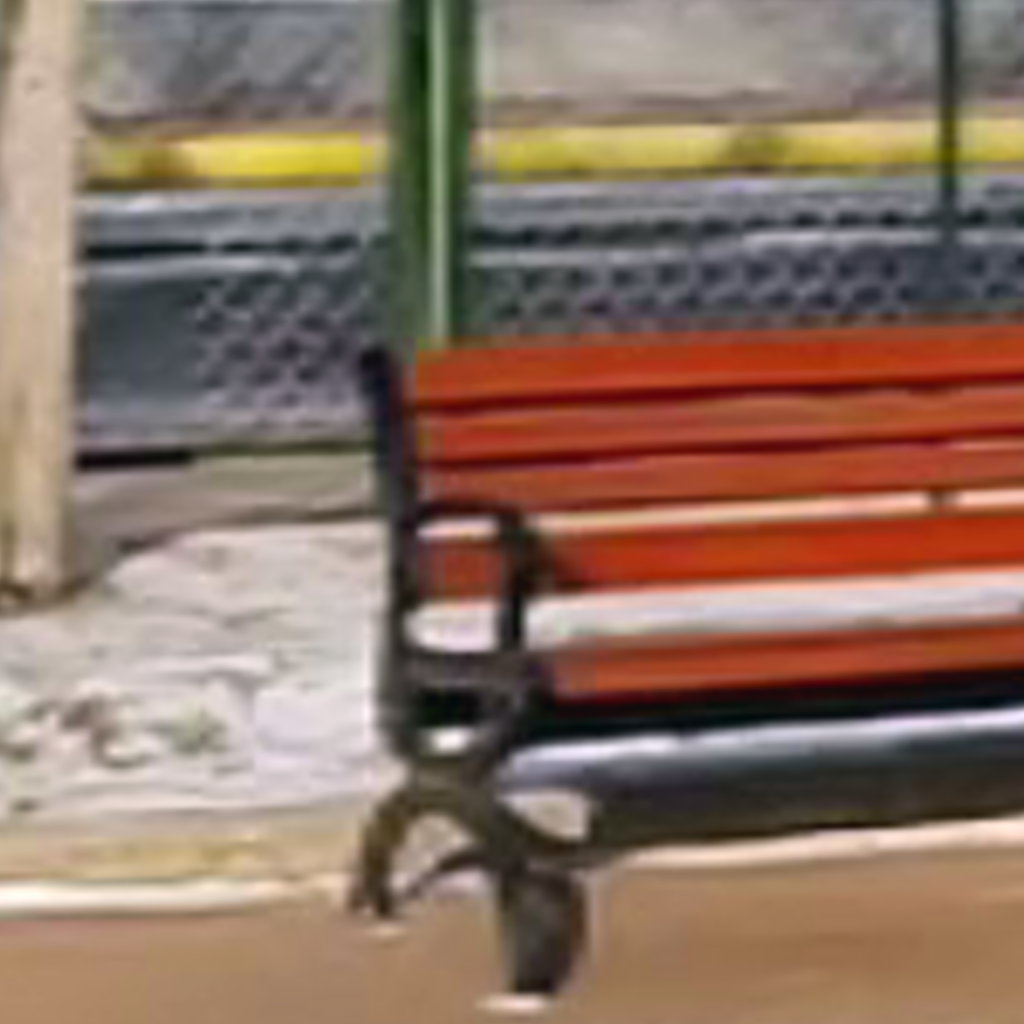}
        \hspace{-1mm}
            \includegraphics[width=.107\linewidth]{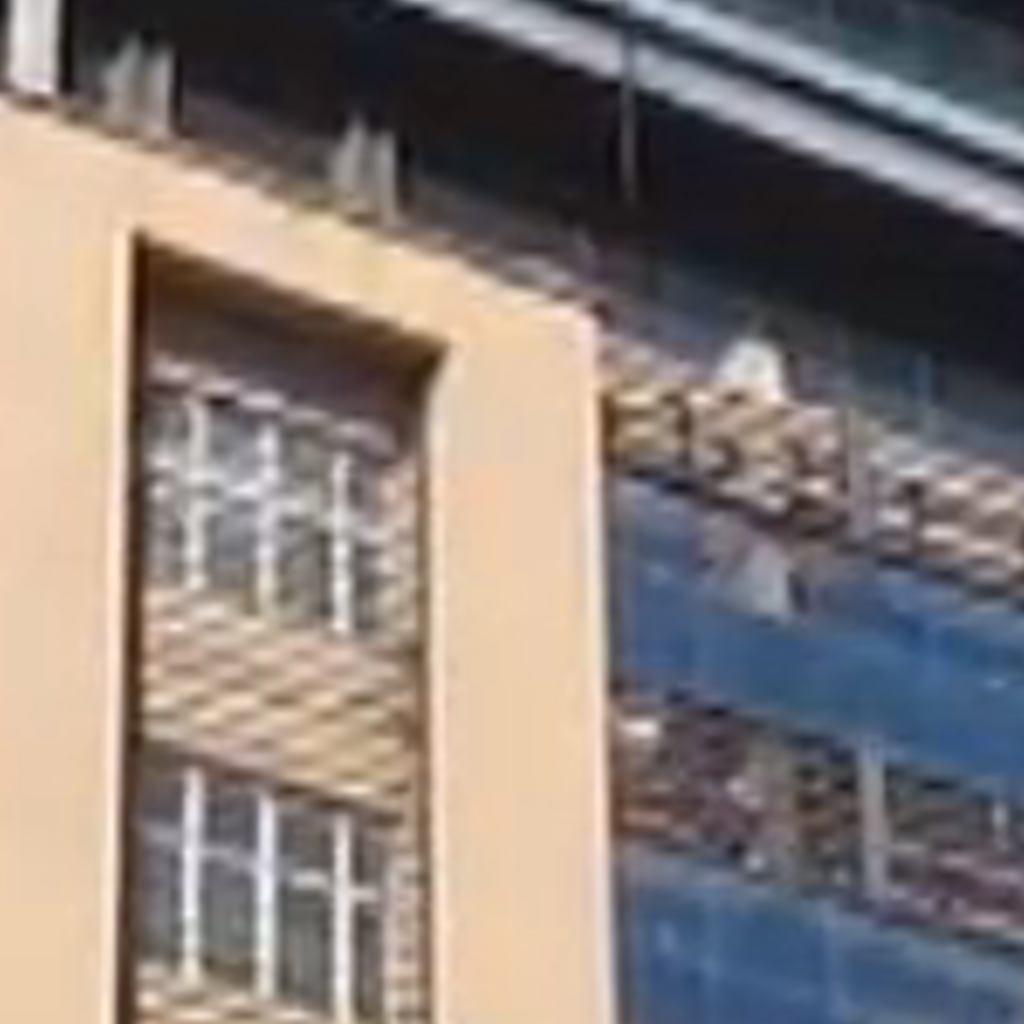}
        }\\\vspace{-4mm}
        \subcaptionbox*{}
        {
            \includegraphics[width=.107\linewidth]{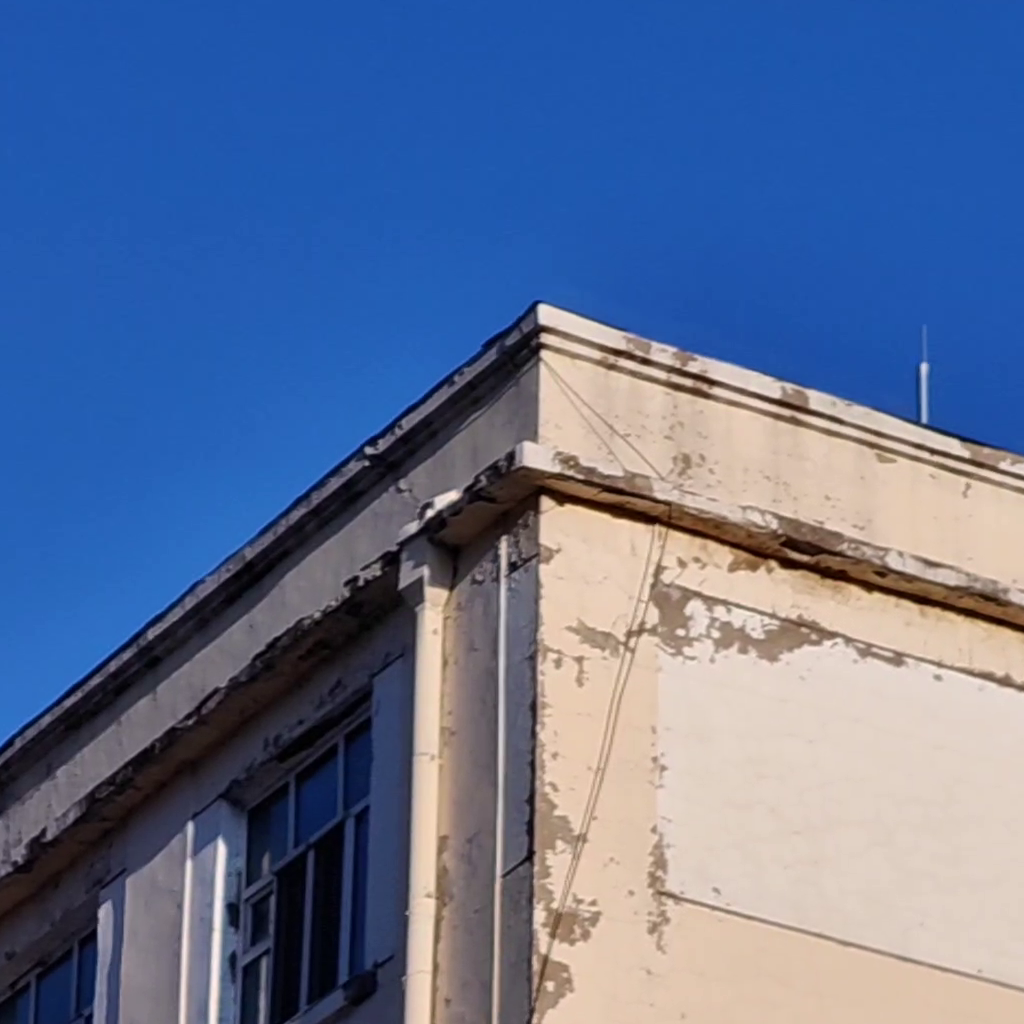}
        \hspace{-1mm} 
            \includegraphics[width=.107\linewidth]{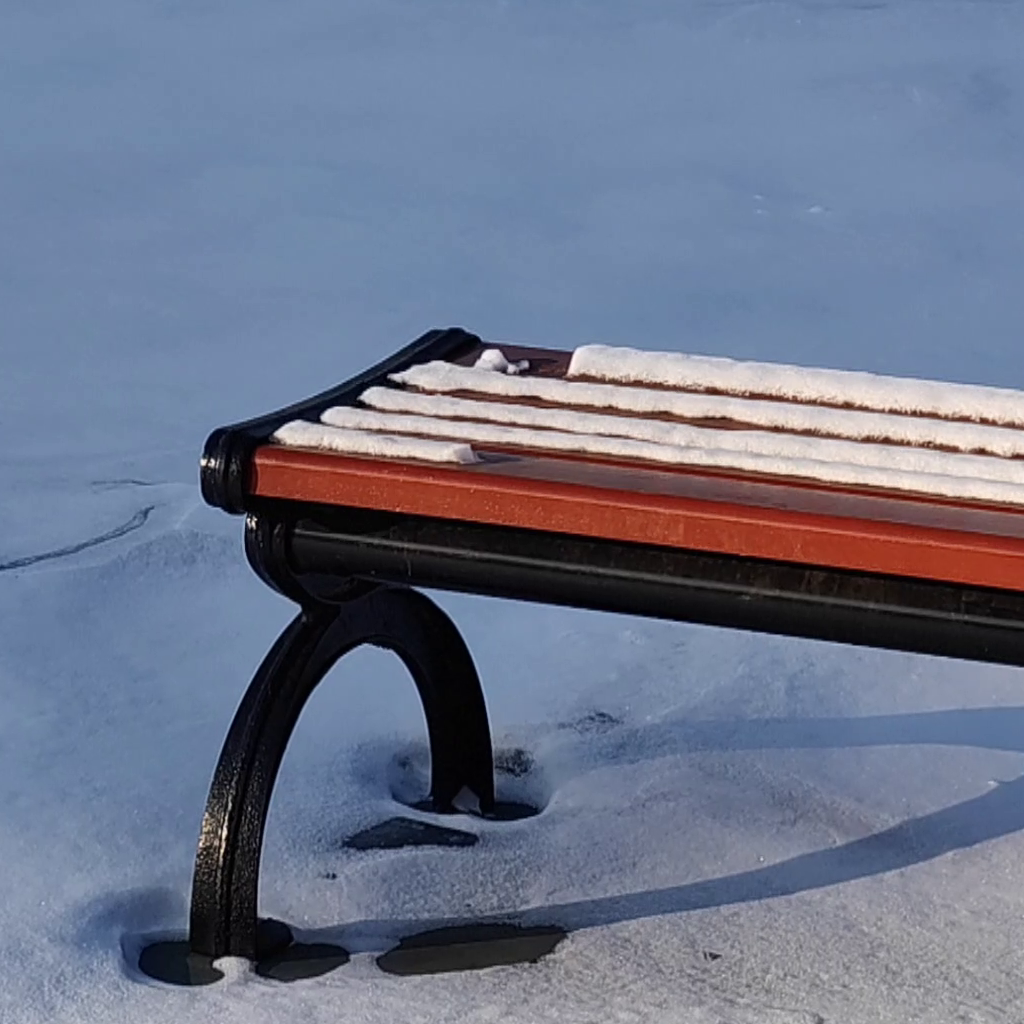}
        \hspace{-1mm}
            \includegraphics[width=.107\linewidth]{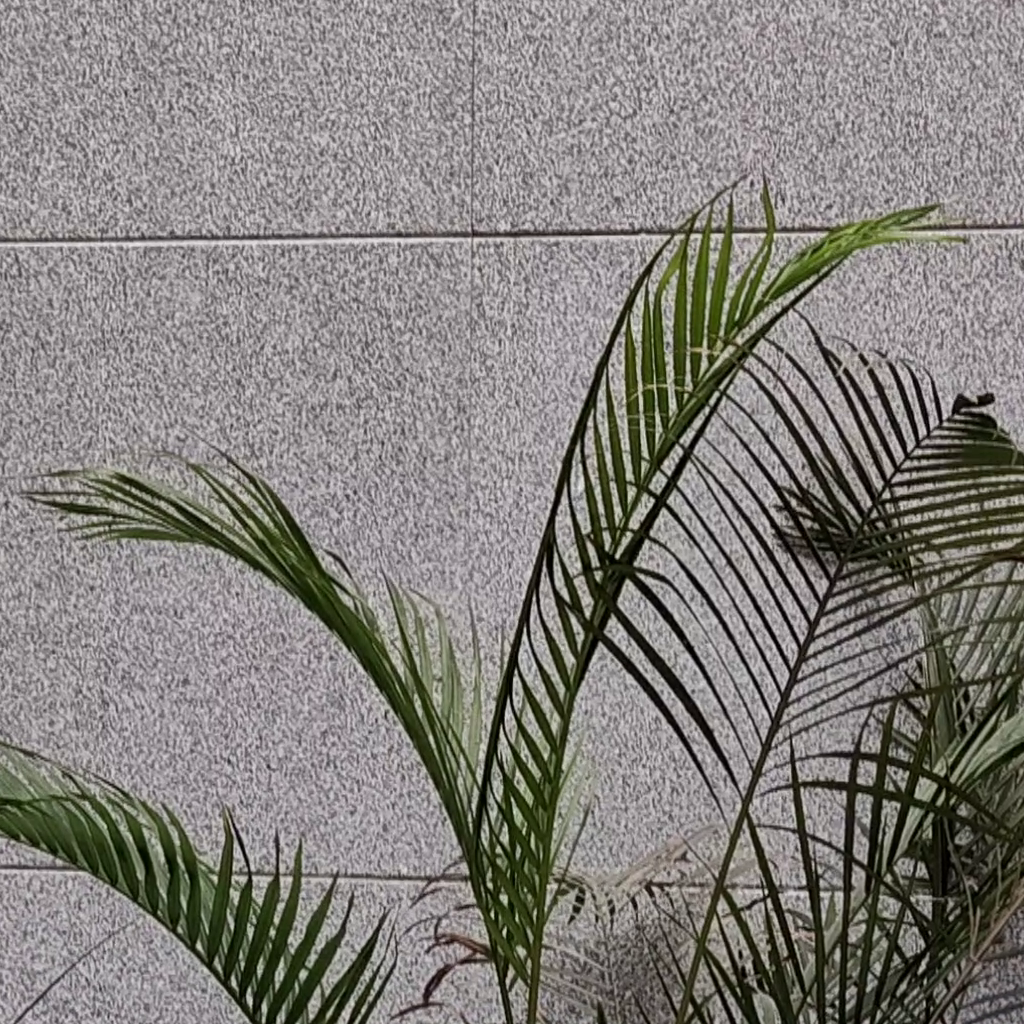}
        \hspace{-1mm}
            \includegraphics[width=.107\linewidth]{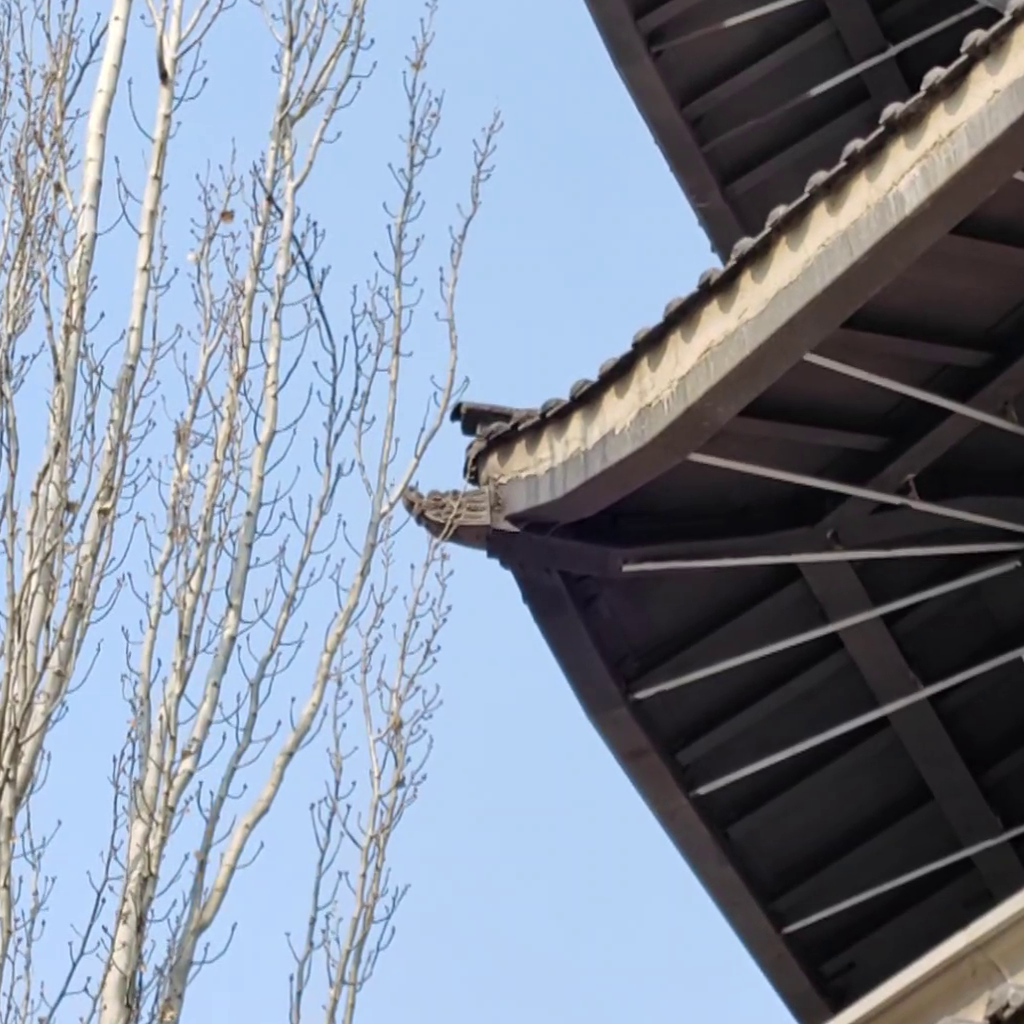}
        \hspace{-1mm}
            \includegraphics[width=.107\linewidth]{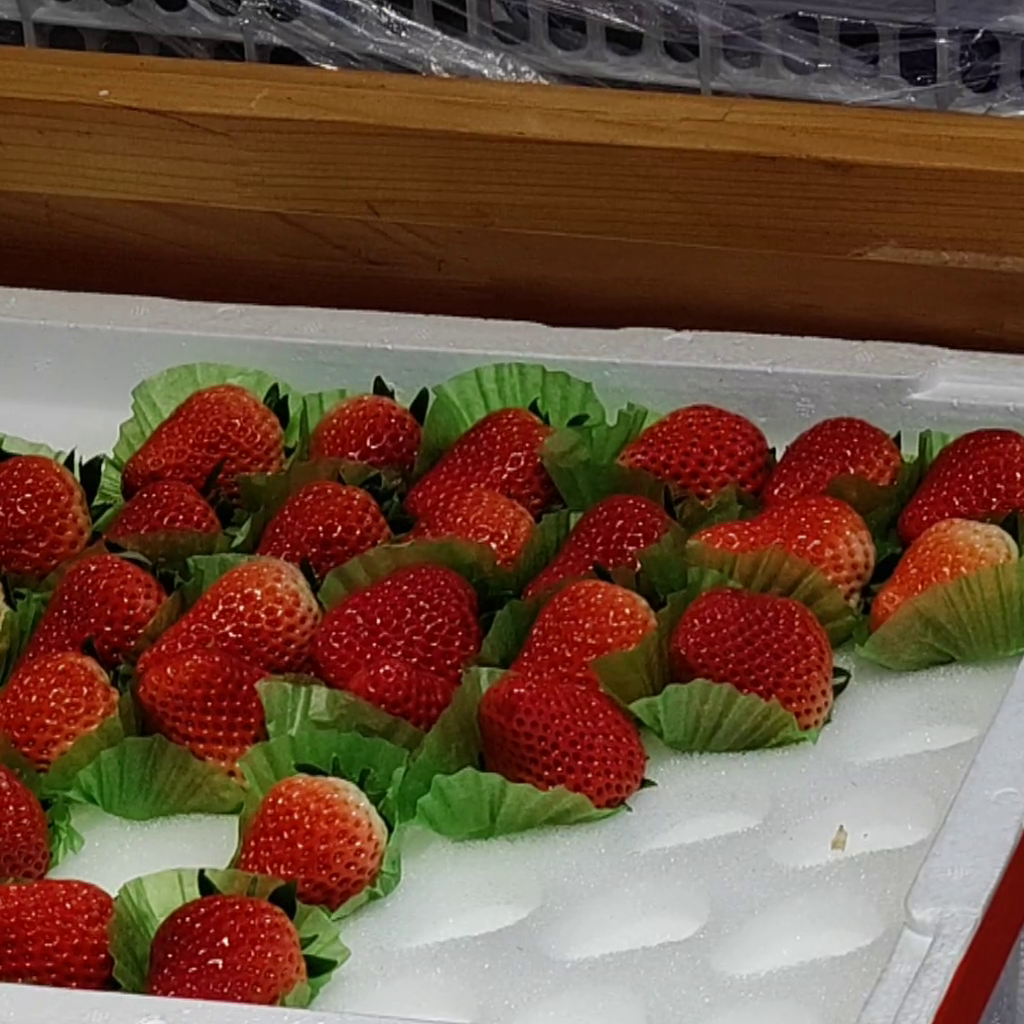}
        \hspace{-1mm}
            \includegraphics[width=.107\linewidth]{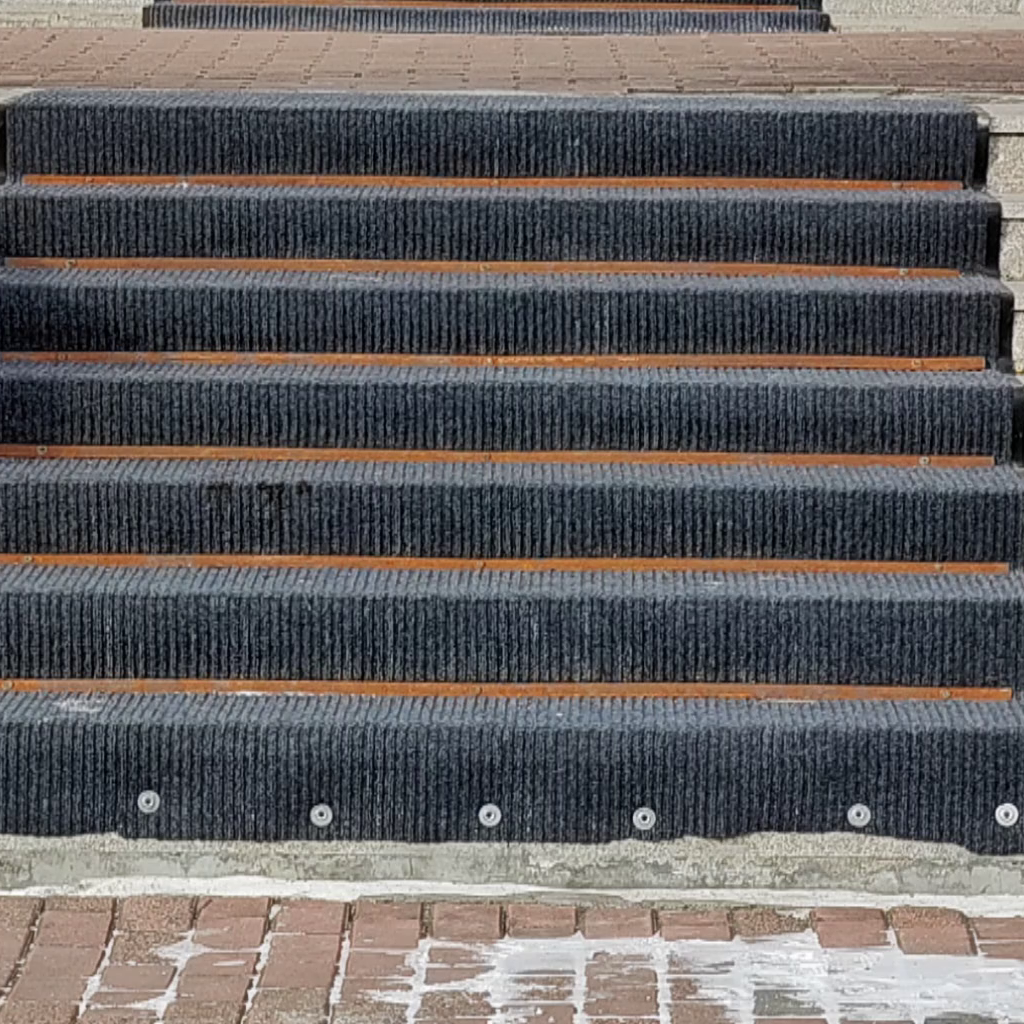}
        \hspace{-1mm}
            \includegraphics[width=.107\linewidth]{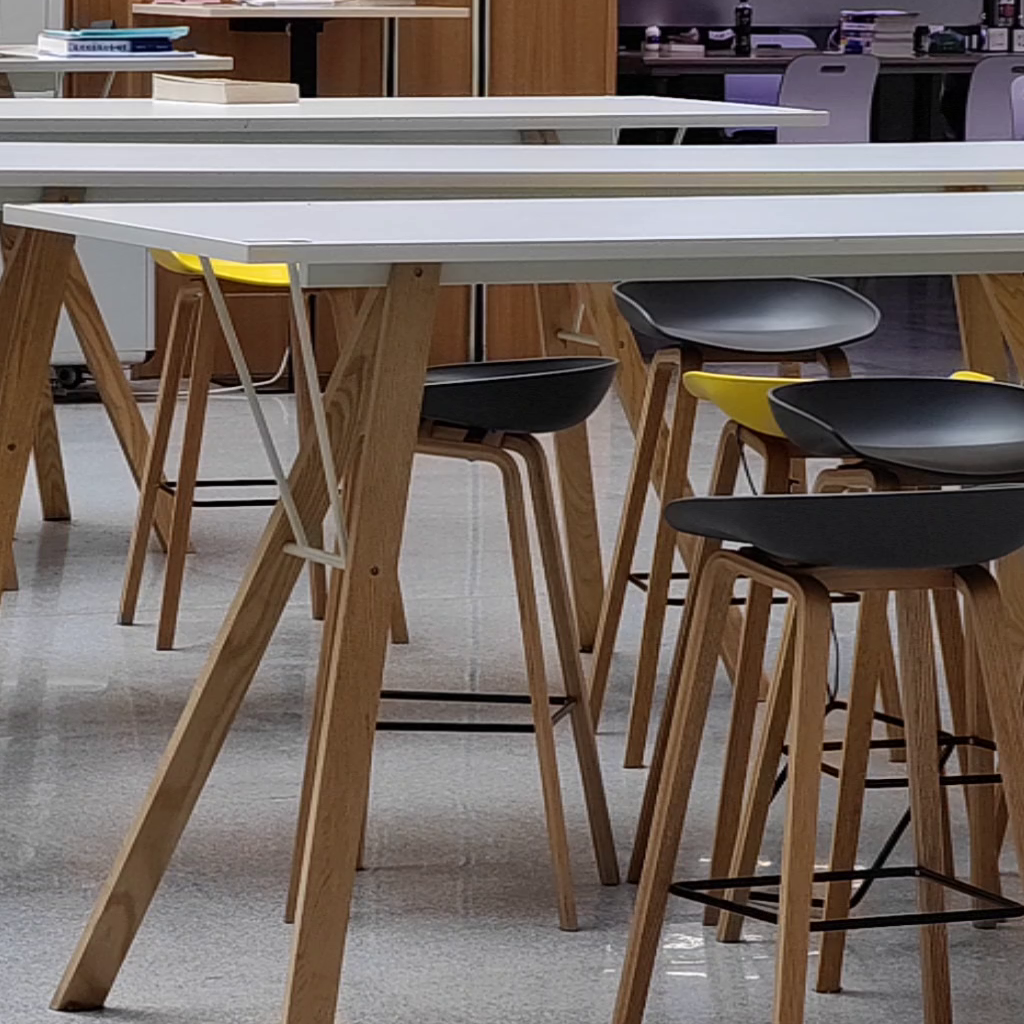}
        \hspace{-1mm}
            \includegraphics[width=.107\linewidth]{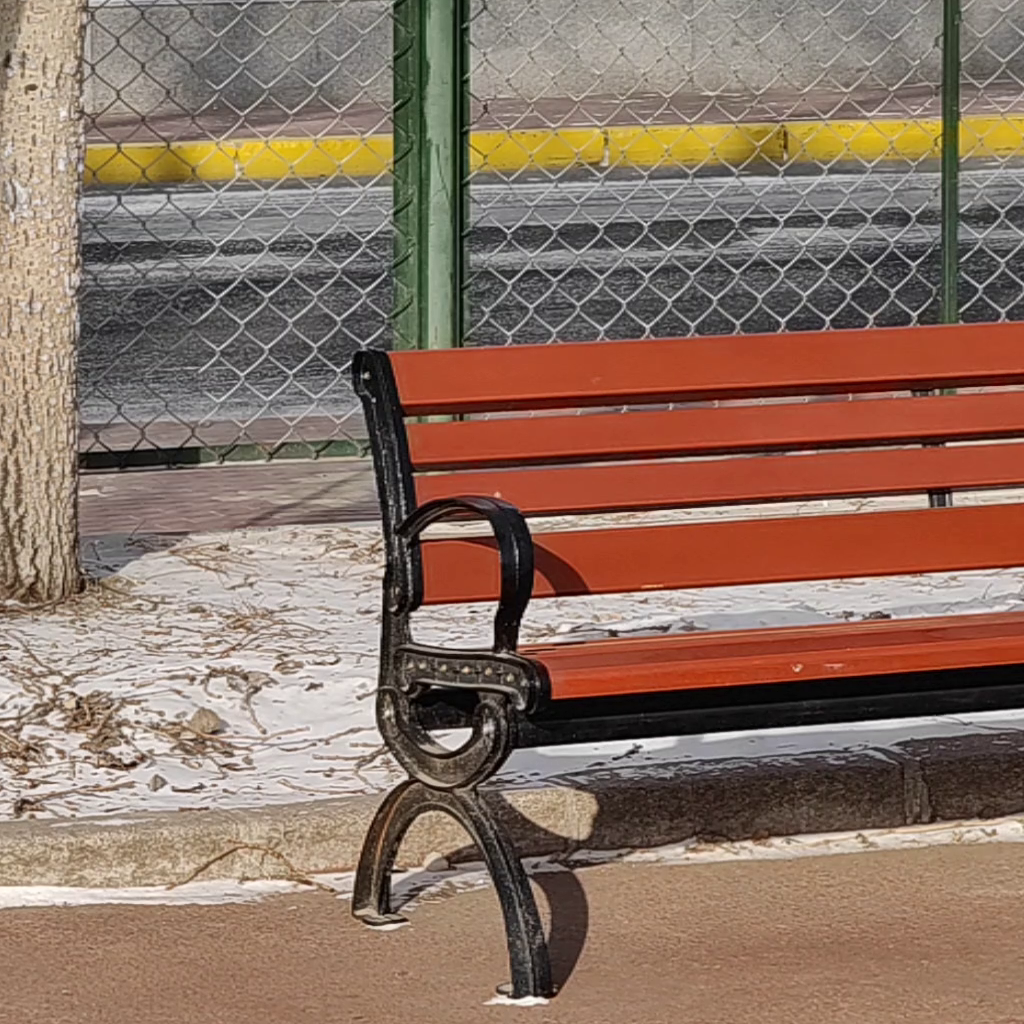}
        \hspace{-1mm}
            \includegraphics[width=.107\linewidth]{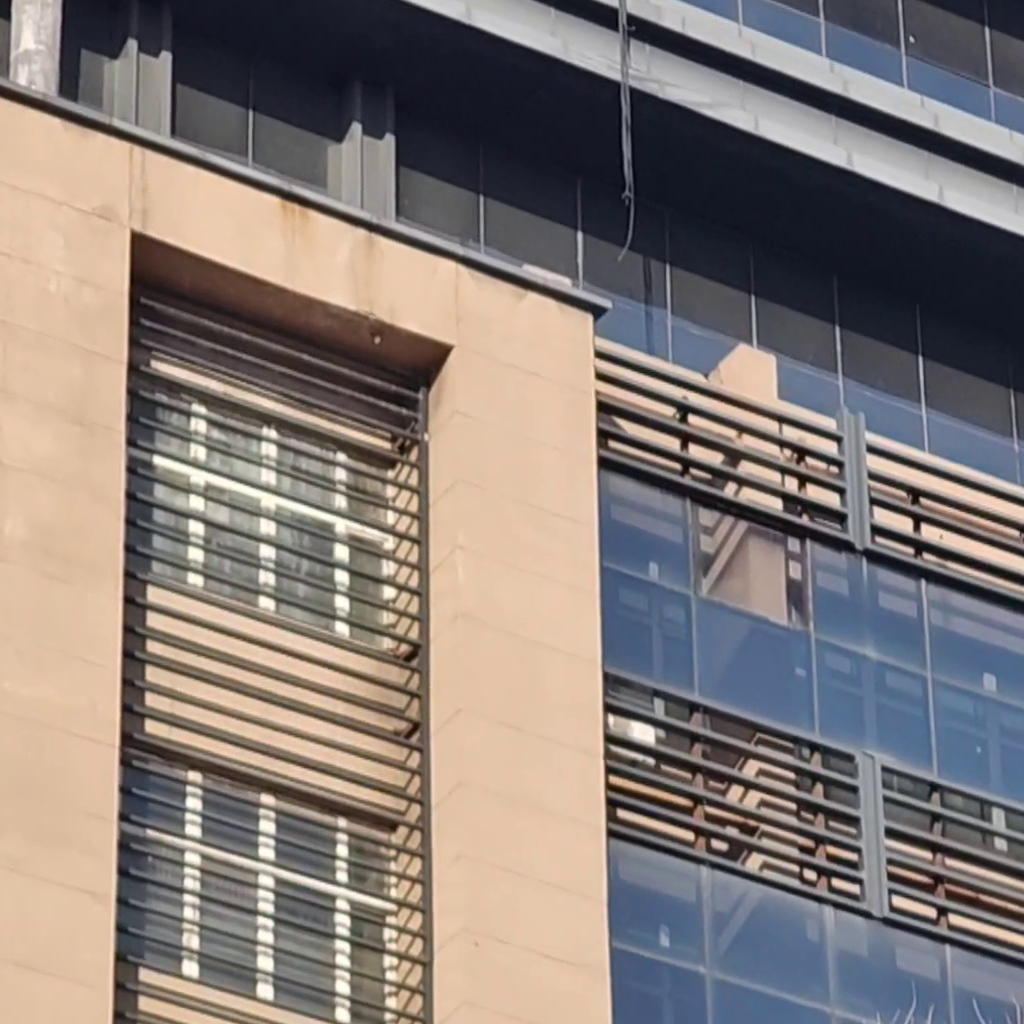}
        }
    \vspace{-3mm}
    \caption{Some images from the MVSR4$\times$ dataset. The first row shows examples of LR frames and the second row shows examples of corresponding aligned HR frames.}
    \label{fig:dataset} % is more photo-realistic.
    \vspace{-2mm}
\end{figure*}
Existing VSR methods are mostly trained on synthetic data where LR sequences are obtained by some simple degradations ({\em e.g.}, bicubic downsampling) of HR sequences.
However, these degradations can not fully reflect the degradation of real-world videos, and the VSR methods hardly generate high-quality results in real scenarios.
A feasible solution is to construct real-world datasets to train VSR models.
Some relevant datasets have been proposed, the comparison can be seen in Table~\ref{tab:dataset}.
For instance, Chan {\em et al.}~\cite{chan2022investigating} collect the real-world VSR dataset VideoLQ, which only has LR sequences but not their HR counterparts.
While RealMCVSR~\cite{Lee_2022_CVPR} captures paired videos by smartphone lenses with different focal lengths, the data pairs are not aligned.
%, and the alignment of large scale factor datasets is more complicated, whose degradation is more complex, and wide-angle distortion is more serious.
%
In addition, Yang {\em et al.}~\cite{yang2021real} proposed a $\times$2 real-world aligned VSR dataset. %, where LR and HR videos are captured by different focal length lenses of mobile phones~\cite{yang2021real}.
However, it is not sufficient to verify the effectiveness of VSR methods, as the scale factor is low and the degradation is a little weak.
Therefore, there is still a high demand for paired, aligned, and larger scale factor VSR datasets from the real world. 

\begin{figure}[t!]
    % \vspace{-3mm}
    \centering
        \centering
        \includegraphics[width=\linewidth]{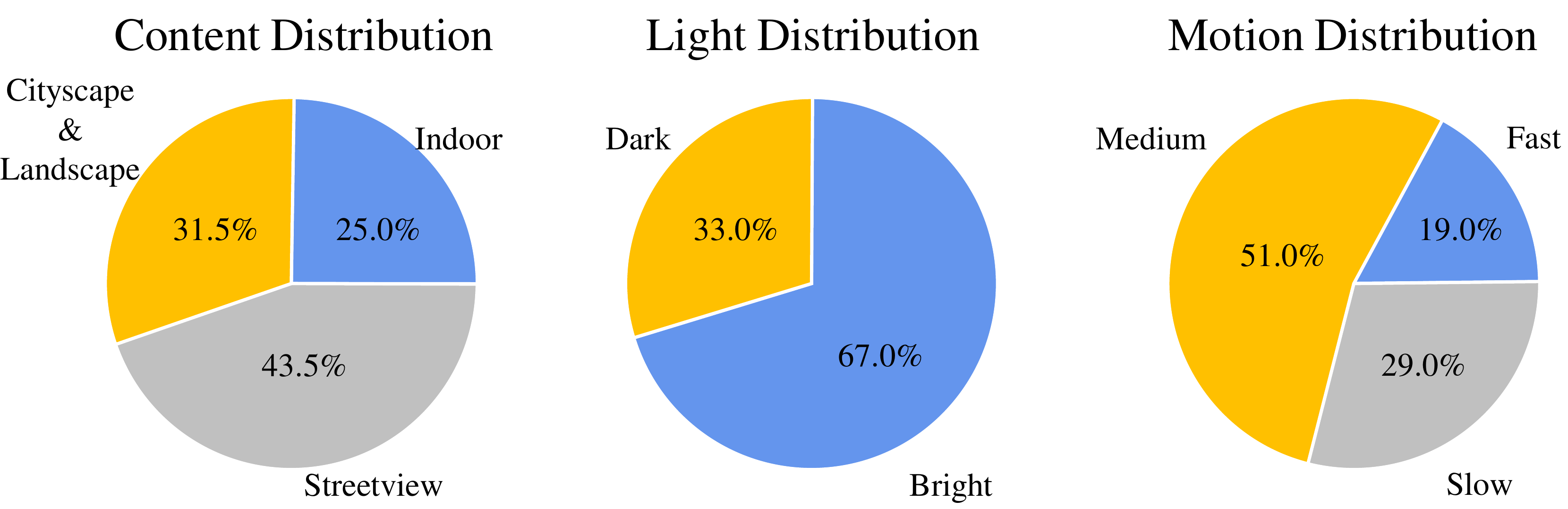}
    % \vspace{-2.5mm}
    \caption{Content and scenarios analysis of our MVSR4$\times$ dataset.}
    \label{fig:content} % is more photo-realistic.
\end{figure}

In this work, we collect a new real-world VSR dataset with the scale factor of 4 by mobile phone ({\em i.e.}, Huawei P50 Pro), named MVSR4$\times$. 
As illustrated in Fig. \ref{fig:process}, we use an ultra-wide camera with a 13mm-equivalent ($r_u$) lens and a telephoto camera with a 90mm-equivalent ($r_t$) lens to capture LR and HR sequences, respectively. 
Thanks to the dual-view video function provided by the camera engine, we can easily capture LR and HR videos at the same frequency and almost synchronously.
The focal length scale of HR and LR videos is $r_t/r_u\approx6.9$.
For simplicity, we regard it as the dataset with a common scale factor of 4.
Thus, we need to crop the LR size to $r_u/(r_t/4)\approx0.58$ times the original size.
This operation can also remove areas with wide-angle distortion. 
Finally, we utilize the alignment algorithm of RealSR \cite{cai2019toward} to correct the spatial position and color of LR in an iterative manner. 
We select 300 high-quality LR-HR sequences from the captured 500 videos to form the MVSR4$\times$ dataset.
Each sequence has 100 frames whose frame rate is 30fps and resolution is 1080P. 
Fig.~\ref{fig:dataset} illustrates some image examples.
Moreover, the dataset consists of diverse contents, including slow and fast movement, bright and dark, indoor and outdoor scenes, as shown in Fig.~\ref{fig:content}.
There are 280 sequences used for training and 5 sequences for validation.
The remaining 15 sequences are used for testing.

\section{Method}
\label{sec:method}

In this section, we first give an overview of our pipeline in Sec.~\ref{seq:overview}.
Then we introduce the proposed alignment method for neighboring frames from the real world in Sec.~\ref{seq:multi} and training loss in Sec.~\ref{seq:loss}.

\subsection{Overview}\label{seq:overview}
The whole VSR pipeline consists of four parts, {\em i.e.}, encoding model, alignment model, reconstruction model, and upsampling model.

Denote by $\textit{x}\in\mathbb{R}^{\textit{T}\times\textit{H}\times\textit{W}\times\textit{C}}$ low-resolution videos. 
When reconstructing the \textit{i}-th frame  ${x_i}$, ${x_i}$ is firstly fed into the encoder model for feature extraction,
\begin{equation}
    \begin{aligned}
        f_i&=\mathcal{E}\left(x_i\right),
    \end{aligned}
\end{equation}
where $\mathcal{E}$ denotes the encoder and $f_i$ is the output features.
Taking forward propagation as an example,
the alignment model calculates the offsets between the current frame $x_i$ and the previous frame $x_{i-1}$, and then warps the reconstruction features $h_{i-1}$ of $x_{i-1}$ to get the aligned features $\bar{h}_{i}$.
The restoration features are recursively generated from the restoration model $\mathcal{R}$, it can be expressed as,
\begin{equation}
% \vspace{-2mm}
    \begin{aligned}
        h_i&=\mathcal{R}\left(x_i,\bar{h}_i\right).\\
    \end{aligned}
    % \vspace{-1mm}
\end{equation}
Finally, the upsampling module $\mathcal{U}$ is deployed to generate the super-resolution result $\hat{y}_i$,
\begin{equation}
    \begin{aligned}
    \hat{y}_i&=\mathcal{U}\left({h}_i\right), \label{rec}
    \end{aligned}
    % \vspace{-1mm}
\end{equation}
where $\hat{y}_i\in\mathbb{R}^{\textit{T}\times\textit{rH}\times\textit{rW}\times\textit{C}}$ and \textit{r} denotes the scale factor of super-resolution.

To estimate the effectiveness of the proposed method comprehensively, we select bidirectional~\cite{chan2021basicvsr} and second-order grid~\cite{chan2021basicvsr++} as our temporal propagation scheme, named EAVSR and EAVSR+, respectively.
Due to the space limitation, the specific structure is introduced in the supplementary material.

\subsection{Inter-Frame Ailgnment} \label{seq:multi}

\begin{figure*}[!htbp]
    \vspace{-3mm}
    \centering
        \centering
        \includegraphics[width=0.90\linewidth]{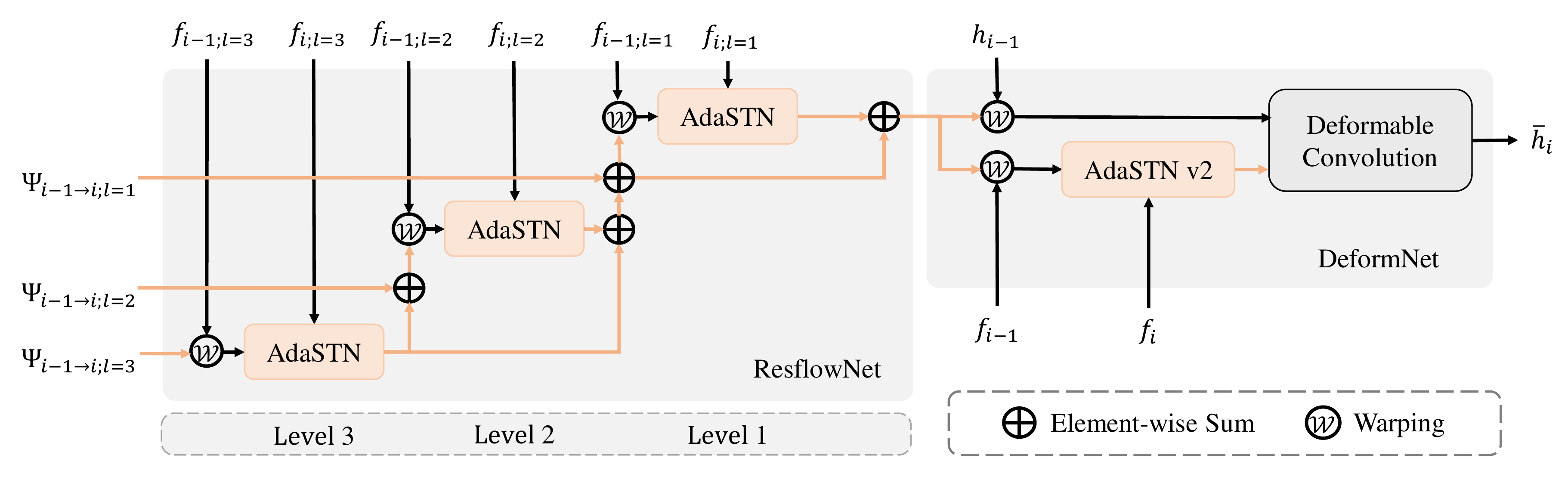}
    \caption{The structure of MultiAdaSTN. Colored lines and black lines represent the transfer direction of offsets and features, respectively.}
    \label{fig:multiadastn} % is more photo-realistic.
    \vspace{-1mm}
\end{figure*}
\begin{figure}[!htbp]
    \vspace{-3mm}
    \centering
        \centering
        \includegraphics[width=0.95\linewidth]{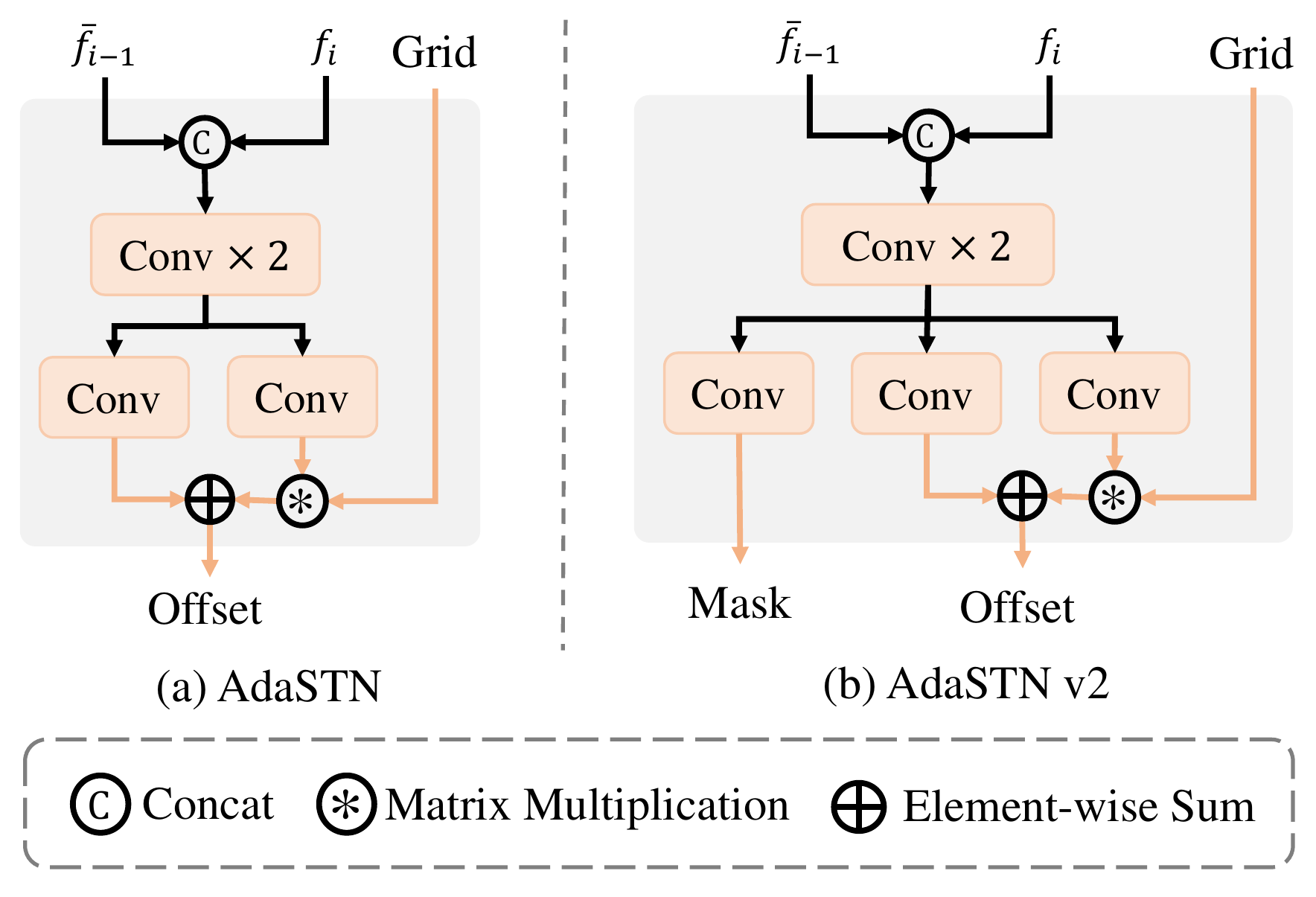}
    \caption{The structure of AdaSTN and AdaSTN v2. Colored lines and black lines represent the transfer direction of offsets and features, respectively.}
    \label{fig:adastn} % is more photo-realistic.
    \vspace{-1mm}
    % \vspace{-6mm}
\end{figure}
Since the pre-trained optical flow network sometimes provides inaccurate results on real-world data and deformable convolution~\cite{dai2017deformable} is unstable during training.
We consider designing a more effective alignment method for real-world VSR.
Specifically, we propose a multi-layer adaptive spatial transform network (MultiAdaSTN) to refine inter-frame offsets provided by a pre-trained optical flow network. 
As shown in Fig.~\ref{fig:multiadastn}, the proposed MultiAdaSTN contains ResflowNet and DeformNet.

\noindent\textbf{ResflowNet.}
Given two neighboring frames $x_{i-1}$ and $x_{i}$, we first calculate a basic optical flow ${\rm\Psi}_{i-1{\to}i}$ from $x_{i-1}$ to $x_i$ by a pre-trained optical flow network  $\mathcal{F}_s$ (\textit{e.g.}, SpyNet~\cite{ranjan2017optical}),
\begin{equation}
% \vspace{-2mm}
   {\rm\Psi}_{{i-1}\to{i}}=\mathcal{F}_{s}\left(x_{i-1},x_i\right).\label{spynet}
%   \vspace{-1mm}
\end{equation}
ResflowNet estimates a residual optical flow ${\rm\Delta\Psi}_{{i-1}\to{i}}$ on the basis of ${\rm\Psi}_{i-1{\to}i}$ in a coarse-to-fine manner.
To generate the multi-level features $f_{i;l}$, $f_{i-1;l}$ and flow ${\rm\Psi}_{i-1\to{i};l}$ at the $l$-th level, the features and flow at the (\textit{l}-1)-th level are downsampled by a factor of 2. 
In this way, we can obtain all \textit{l}-level features and flow.

Before performing the \textit{l}-th level, the residual flow ${\rm\Delta\Psi}_{{i-1}\to{i;l+1}}$ from (\textit{l}+1)-th level can be acquired.
Then the residual and basic flow are added together to generate the \textit{l}-th coarse flow,
\begin{equation}
   {\rm\Psi}_{{i-1}\to{i;l}}={\rm\Psi}_{i-1{\to}i;l}+\left({\rm\Delta\Psi}_{{i-1}\to{i;l+1}}\right)_{\uparrow2},\label{first}
\end{equation}
where $\left(\cdot\right)_{\uparrow{s}}$ refers to the bilinear upsampling with the scale factor of \textit{s}. Next, $f_{i-1;l}$ can be warped with ${\rm\Psi}_{{i-1}\to{i;l}}$ to obtain the preliminary aligned feature $\bar{f}_{i-1;l}$, 
\begin{equation}
    \bar{f}_{i-1;l}=\mathcal{W}\left(f_{i-1;l},{\rm\Psi}_{{i-1}\to{i;l}}\right),
\end{equation}
where $\mathcal{W}$ denotes warping operation. 
Following, we utilize the AdaSTN~\cite{zhang2022selfsupervised} $\mathcal{S}$ (shown in Fig.~\ref{fig:adastn} (a)) to calculate fine residual flow between $\bar{f}_{i-1;l}$ and $f_{i;l}$,
\begin{equation}
% \vspace{-1mm}
   {\rm\Delta\tilde\Psi}_{{i-1}\to{i;l}}=\mathcal{S}\left(f_{i;l},\bar{f}_{i-1;l}\right).
%   \vspace{-1mm}
\end{equation}
And the residual flow ${\rm\Delta\Psi}_{{i-1}\to{i};l}$ between $f_{i;l}$ and $f_{i-1;l}$ can be regarded as the sum of residue propagated from (\textit{l}+1)-th level and calculated in \textit{l}-th level,
\begin{equation}
% \vspace{-1mm}
    {\rm\Delta\Psi}_{{i-1}\to{i};l}=\left({\rm\Delta\Psi}_{{i-1}\to{i;l+1}}\right)_{\uparrow2}+{\rm\Delta\tilde\Psi}_{{i-1}\to{i;l}}.\label{last}
    % \vspace{-1mm}
\end{equation}

We loop from Eqn. \ref{first} to Eqn. \ref{last} a total of \textit{l} times and use the residual flow generated in the 1st-level as the ultimate residual flow ${\rm\Delta\Psi}_{{i-1}\to{i}}$, \ie,
\begin{equation}
% \vspace{-1mm}
    {\rm\Delta\Psi}_{{i-1}\to{i}}={\rm\Delta\Psi}_{{i-1}\to{i};l=1}.
    % \vspace{-1mm}
\end{equation}
So far, we can obtain a more precise and task-oriented optical flow estimation.
Then we present DeformNet to further refine inter-frame alignment, where modified AdaSTN (\ie, AdaSTN v2) is utilized to predict the offsets and masks of deformable convolution v2~\cite{zhu2019deformable}.

\noindent\textbf{DeformNet.}
AdaSTN~\cite{zhang2022selfsupervised} predicts the pixel-level affine transformation matrix and translation vector between two misalignment features.
However, it predicts the same offset for all convolutional channels and ignores the out-of-bounds issues, which reduces its flexibility. 
Instead, we present AdaSTN v2 to improve it, as shown in Fig.~\ref{fig:adastn} (b).
The predicted offset for each pixel of AdaSTN v2 can be written as,
\begin{equation}
    \mathbf{P}=\mathbf{AG}+\mathbf{b},
\end{equation}
where $\mathbf{A}\in\mathbb{R}^{n{\times}2\times2}$ is the estimated affine transformation matrix, $\mathbf{b}\in\mathbb{R}^{n{\times}2\times1}$ is the translation vector, and $n$ is the number of convolutional groups. 
$\mathbf{G}$ is a positional grid, which can be expressed as,
\begin{equation}
    \mathbf{G}=\left[\begin{array}{rrrrrrrrr} 
    -1 & -1 & -1 & 0 & 0 & 0 & 1 & 1 & 1\\ 
     -1 & 0 & 1 & -1 & 0 & 1 & -1 & 0 & 1\end{array}\right].
\end{equation}
In addition, we add another branch to calculate masks to block out out-of-bounds and lousy information.

AdaSTN v2 can be also considered as a variant of STN~\cite{jaderberg2015spatial}, and it has the potential to estimate more stable offsets than deformable convolution~\cite{dai2017deformable}.
In DeformNet, we first warp $f_{i-1}$ and  $h_{i-1}$ to obtain aligned features by optical flow ${\rm\Psi}_{{i-1}\to{i}}$, then use an AdaSTN v2 to acquire the deformable offsets  $\psi_{i-1\to{i}}$ and masks $m_{i-1\to{i}}$ between these two warped features,
\begin{align}
% \vspace{-1mm}
        \psi_{i-1\to{i}}, m_{i-1\to{i}}&=\mathcal{S}_{v2}\left(f_i,\mathcal{W}\left(f_{i-1},{\rm\Psi}_{{i-1}\to{i}}\right)\right). \label{hata} 
        % \vspace{-1mm}
\end{align}
Finally, we deploy a deformable convolution $\mathcal{DC}$ to get the final aligned features $\bar{h}_{i}$,
\begin{align}
        \bar{h}_{i}&=\mathcal{DC}\left(\mathcal{W}\left(h_{i-1},{\rm\Psi}_{{i-1}\to{i}}\right), \psi_{i-1\to{i}}, m_{i-1\to{i}}\right). \label{hatb} 
\end{align}
% Noted that Eqn. \ref{hata} and Eqn. \ref{hatb} are the decomposition of Eqn. \ref{hat}.

\subsection{Training Loss} \label{seq:loss}

Despite data pre-processing deployed on both our MVSR4$\times$ dataset and RealVSR~\cite{yang2021real} dataset, spatial position misalignment and color inconsistency still exist between LR-HR pairs.
Learning with these data pairs will lead to pixel shifts and blurry results~\cite{Zhang_2021_ICCV}.
To alleviate the adverse effect of position and color difference, we further utilize the PWC-Net~\cite{sun2018pwc} and the guided filtering~\cite{he2012guided} at the patch level during training, respectively.

Denote by $x_i$ the low-resolution frame and $y_i$ its high-resolution counterpart, we first use guided filtering to transfer the color of $y_i$ to close to $x_i$,
\begin{equation}
    y_i^{g}=\mathcal{G}\left((x_i)_{\uparrow{r}},y_i\right),
\end{equation}
where $\mathcal{G}$ denotes guided filtering operation and $y_i^{g}$ denotes the ground truth after converting color.
Then we apply a pre-trained optical flow network $\mathcal{F}$ (\ie, PWC-Net~\cite{sun2018pwc}) to estimate the optical flow $\mathcal{O}$ from $y_i^{g}$ to $x_i$,
\begin{equation}
    \mathcal{O}=\mathcal{F}\left(x_i,(y_i^{g})_{\downarrow{r}}\right).
\end{equation}
The upsampled optical flow $(\mathcal{O})_{\uparrow{r}}$ is used to warp $y_i^{g}$,
\begin{equation}
    y^w=\mathcal{W}\left(y_i^{g}, \left(\mathcal{O}\right)_{\uparrow{r}}\right),
\end{equation}
where $y^w$ is a well-aligned target HR patch for supervising the VSR network.
The loss for training our VSR network can be written as,
\begin{equation}
\mathcal{L}\left(\hat{y}_i,y^w\right)=\Vert{\mathbf{m}}\circ\left(\hat{y}_i-y^w\right)\Vert_1,
\end{equation}
where $\circ$ means the pixel-wise product, and $\mathbf{m}$ is the out-of-bounds mask calculated by the optical flow $\mathcal{O}$.

%------------------------------------------------------------------------
\section{Experiments}
\label{sec:exp}
\subsection{Implementation Details}
Experiments are conducted on real-world VSR datasets RealVSR \cite{yang2021real} and MVSR4$\times$ for training and evaluation.

\noindent\textbf{Datasets.} The RealVSR dataset has 450 videos for training and 50 videos for inference, and our MVSR4$\times$ randomly selects 280 videos for training and 5 videos for evaluation and leaves the remaining 15 videos for testing.
The details of MVSR4$\times$ dataset can be seen in Sec. \ref{sec:dataset}.

\noindent\textbf{Training Configurations.}
We randomly crop patches of size $128\times128$ from the video frames with the mini-batch size 8 during training.
Data augmentation is applied on training images, including random horizontal flip, vertical flip and $90^{\circ}$ rotation. We adopt Adam optimizer \cite{kingma2014adam} with $\beta_1$ = 0.9 and $\beta_2$ = 0.999 to train the model for 400 epochs. The learning rate is initially set to $1 \times10^{-4}$ and the model is optimized in cosine annealing scheme \cite{loshchilov2016sgdr}.
The experiments are conducted with PyTorch \cite{paszke2019pytorch} framework on an Nvidia GeForce A6000 GPU.

\noindent\textbf{Evaluation Configurations.}
We conduct experiments by comparing EAVSR+ and EAVSR with 11 models: RCAN \cite{zhang2018image}, FSTRN \cite{li2019fast}, TOF \cite{xue2019video}, TDAN \cite{tian2020temporally}, EDVR \cite{wang2019edvr}, RSDN \cite{isobe2020video}, BasicVSR \cite{chan2021basicvsr}, ETDM~\cite{Isobe_2022_CVPR}, MANA~\cite{MANA}, TTVSR~\cite{Liu_2022_CVPR}, and BasicVSR++ \cite{chan2021basicvsr++}. All of these methods are retrained with $\ell_1$ loss on  RealVSR and MVSR4$\times$ datasets, respectively.
% with the official codes and settings in RGB-channel.
% 
Moreover, we train a model named EAVSRGAN+ with the perceptual loss~\cite{johnson2016perceptual} and adversarial loss~\cite{creswell2018generative}, which is the same as the loss term of RealBasicVSR~\cite{chan2022investigating}. 
And the official model of RealBasicVSR is used for testing on RealVSR and MVSR4$\times$ datasets. 

To evaluate the performance quantitatively, we compute three metrics on RGB channels, \textit{i.e.}, Peak Signal to Noise Ratio (PSNR), Structural Similarity (SSIM)~\cite{wang2004image}, and Learned Perceptual Image Patch Similarity (LPIPS)~\cite{zhang2018unreasonable}, where the version of LPIPS is 0.1 trained on the AlexNet~\cite{krizhevsky2017imagenet} network. 

\begin{figure*}[t]
    \centering
    \includegraphics[width=\linewidth]{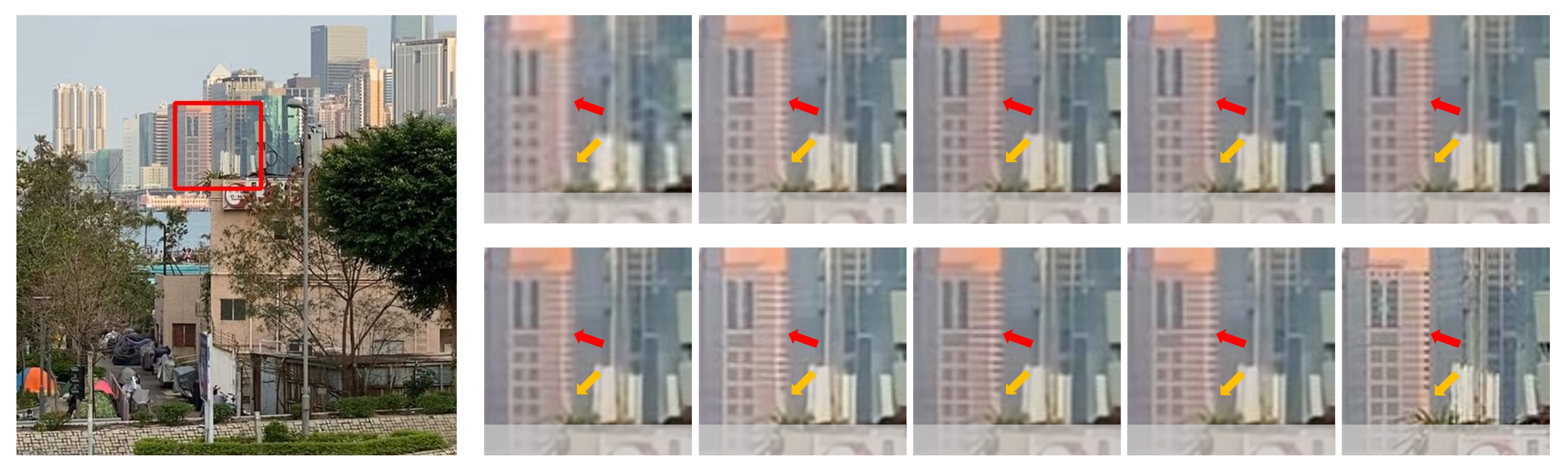}
    \put(-316, 79){\small{LR}}
    \put(-263, 79.5){\small{EDVR~\cite{wang2019edvr}}}
    \put(-196, 79.5){\small{MANA~\cite{MANA}}}
    \put(-132, 79.5){\small{BasicVSR~\cite{chan2021basicvsr}}}
    \put(-60, 79.5){\small{TTVSR~\cite{Liu_2022_CVPR}}}
    \put(-330, 6){\small{ETDM~\cite{Isobe_2022_CVPR}}}
    \put(-272.5, 6){\small{BasicVSR++~\cite{chan2021basicvsr++}}}
    \put(-189, 6){\small{EAVSR}}
    \put(-124, 6){\small{EAVSR+}}
    \put(-45, 6){\small{HR}}
    \caption{Visual comparison on RealVSR dataset~\cite{yang2021real}. Due to the limitation of space, here we select six methods with better PSNR. Our methods EAVSR+ and EAVSR can generate more textures, especially on the building. Please zoom in for details.}
    \label{fig:real-result}
    % \vspace{-4mm}
\end{figure*}

\begin{figure*}[t]
    \centering
    \includegraphics[width=\linewidth]{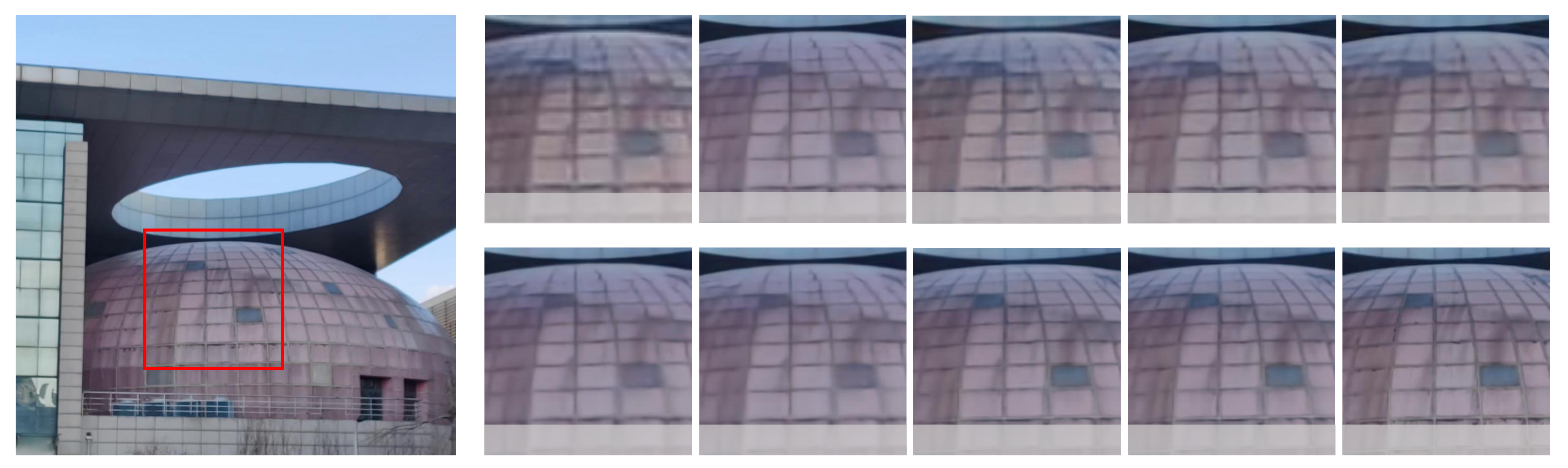}
    \put(-316, 79){\small{LR}}
    \put(-263, 79.5){\small{EDVR~\cite{wang2019edvr}}}
    \put(-196, 79.5){\small{MANA~\cite{MANA}}}
    \put(-132, 79.5){\small{BasicVSR~\cite{chan2021basicvsr}}}
    \put(-60, 79.5){\small{TTVSR~\cite{Liu_2022_CVPR}}}
    \put(-330, 6){\small{ETDM~\cite{Isobe_2022_CVPR}}}
    \put(-272.5, 6){\small{BasicVSR++~\cite{chan2021basicvsr++}}}
    \put(-189, 6){\small{EAVSR}}
    \put(-124, 6){\small{EAVSR+}}
    \put(-45, 6){\small{HR}}
    \caption{Visual comparison on our MVSR4$\times$ dataset. Due to the limitation of space, here we select six methods with better PSNR. Our methods EAVSR+ and EAVSR can restore sharper edges. Please zoom in for details.}
    \label{fig:mvsr4x}
\end{figure*}
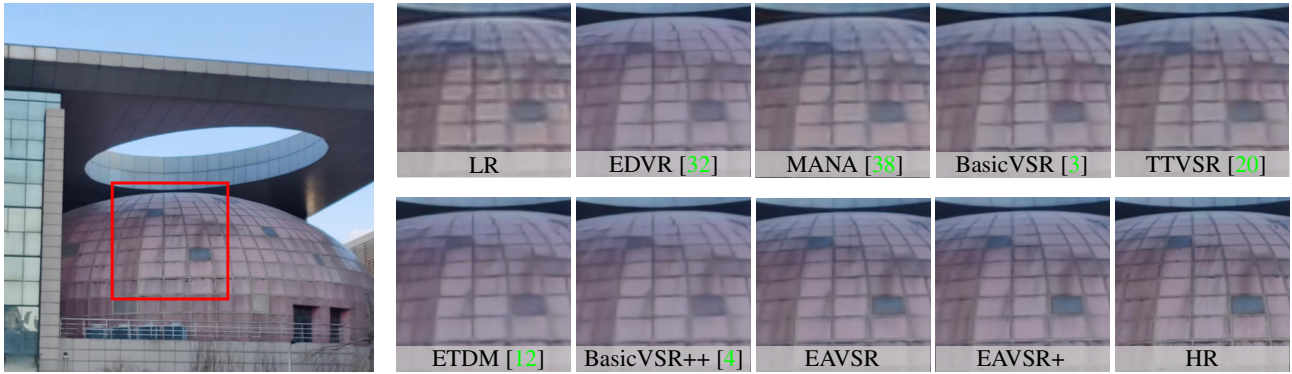

\begin{table*}[t!]
  \small
  \caption{Quantitative comparison on RealVSR dataset~\cite{yang2021real} and our MVSR4$\times$ datasets. \#Frame indicates the number of input frames during inference. `$p$' and `$a$' indicate that the previous frames and all video sequences are required, respectively. \textbf{Bold} marks the best results.}
  \vspace{-2mm}
  \label{tab:result}
  \centering\noindent
  \centering%
  \begin{center}
    \begin{tabular}{clccc}
      \toprule
      Loss Term & Method & \#Frame 
      & \tabincell{c}{{RealVSR~\cite{yang2021real}} \\ {PSNR$\uparrow$ / SSIM$\uparrow$ / LPIPS$\downarrow$}}
      & \tabincell{c}{{MVSR4$\times$} \\ {PSNR$\uparrow$ / SSIM$\uparrow$ / LPIPS$\downarrow$}} \\
        \midrule
        \multirow{13}{*}{w/o Adversarial Loss}
        &RCAN~\cite{zhang2018image} & 1 & 23.59 / 0.7720 / 0.230 &22.72 / 0.7426 / 0.296\\
        &RSDN~\cite{isobe2020video}& $p$ & 23.91 / 0.7743 / 0.224 & 23.15 / 0.7533 / 0.279\\
        &FSTRN~\cite{li2019fast} & 7   &   23.36 / 0.7683 / 0.240 &  22.66 / 0.7433 / 0.315 \\
        &TOF~\cite{xue2019video} & 7   &    23.62 / 0.7739 / 0.220 &   22.80 / 0.7502 / 0.279 \\
        &TDAN~\cite{tian2020temporally} & 7    &   23.71 / 0.7737 / 0.229 &23.07 / 0.7492 / 0.282\\
        &EDVR~\cite{wang2019edvr} & 7  & 23.96 / 0.7781 / 0.216  &23.51 / 0.7611 / 0.268\\
        %  \hline
        &BasicVSR~\cite{chan2021basicvsr} & $a$ & 24.00 / 0.7801 / 0.209 &23.38 / 0.7594 / 0.270\\
        &MANA~\cite{MANA} & $a$ & 23.89 / 0.7781 / 0.224 &23.15 / 0.7513 / 0.285\\
        &TTVSR~\cite{Liu_2022_CVPR} & $a$ & 24.08 / 0.7837 / 0.213 & 23.60 / 0.7686 / 0.277\\
        &ETDM~\cite{Isobe_2022_CVPR} & $a$  & 24.13 / 0.7896 / \textbf{0.206} &23.61 / 0.7662 / 0.260\\
        &BasicVSR++~\cite{chan2021basicvsr++} & $a$ & 24.24 / 0.7933 / 0.216 & 23.70 / 0.7713 / 0.263\\
        &EAVSR (Ours) & $a$  &24.20 / 0.7862 / 0.208
        &23.61 / 0.7618 / 0.264\\
        &EAVSR+ (Ours) & $a$  & \textbf{24.41} / \textbf{0.7953} / 0.212 &\textbf{23.94} / \textbf{0.7726} / \textbf{0.259}\\
        \midrule
        \multirow{2}{*}{w/ Adversarial Loss}\textit{}
        &RealBasicVSR~\cite{chan2022investigating} & $a$ &23.74 / 0.7676 / 0.174 & 23.15 / 0.7603 / 0.202\\
        &EAVSRGAN+ (Ours) & $a$  & \textbf{23.99} / \textbf{0.7726} / \textbf{0.170} & \textbf{23.35} /\textbf{ 0.7611} / \textbf{0.199}\\
      \bottomrule
    \end{tabular}
    \end{center}
    \vspace{-4mm}
\end{table*}

\subsection{Experimental Results on RealVSR}

\begin{figure}[t]
% \vspace{-6mm}
    \centering
    \includegraphics[width=\linewidth]{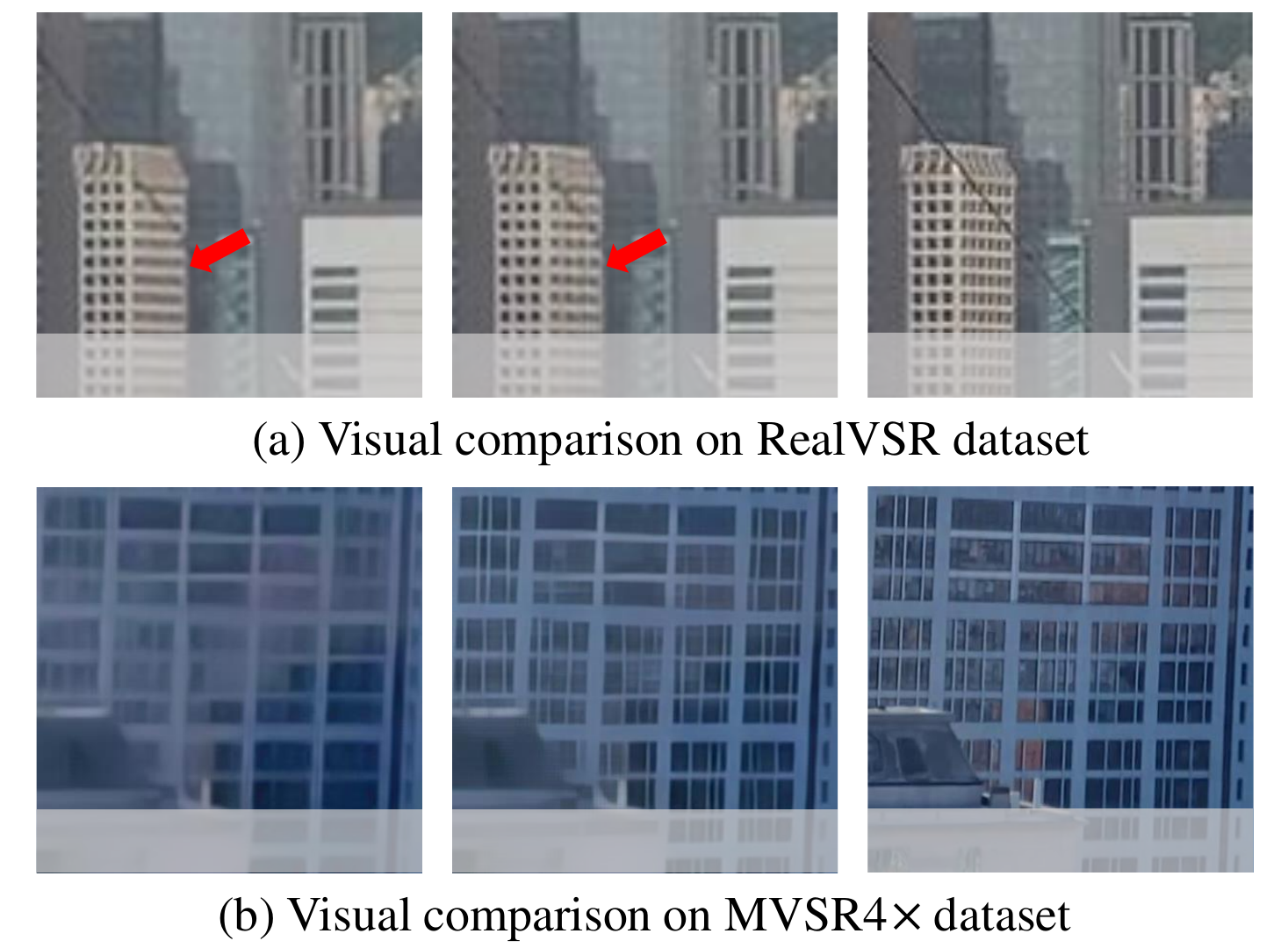}
    % \vspace{-4mm}
    \put(-228, 104){\small{RealBasicVSR~\cite{chan2022investigating}}}
    \put(-134, 104){\small{EAVSR+}}
    \put(-48, 104){\small{HR}}
    \put(-228, 17){\small{RealBasicVSR~\cite{chan2022investigating}}}
    \put(-134, 17){\small{EAVSR+}}
    \put(-48, 17){\small{HR}}
    \caption{Visual comparison between RealBasicVSR~\cite{chan2022investigating} and our EAVSRGAN+ on RealVSR~\cite{yang2021real} and MVSR4$\times$ dataset. Please zoom in for details}
    \label{fig:gan} % is more
    % \vspace{-2mm}
\end{figure}

As shown in Table~\ref{tab:result}, our method EAVSR+ has state-of-the-art performance between all algorithms on RealVSR \cite{yang2021real} dataset. 
EAVSR only changes the alignment strategy and uses the same propagation scheme (bidirectional propagation) and reconstruction module as BasicVSR~\cite{chan2021basicvsr}, and has 0.19dB PSNR gain than BasicVSR.

Besides, we show the qualitative results in Fig. \ref{fig:real-result}. It can be seen that other methods cannot recover some textures on the building well. In contrast, our EAVSR+ and EAVSR can restore clearer texture details with less distortion. 
Furthermore, Fig.~\ref{fig:gan} (a) shows that our EAVSRGAN+ generates more pleasure results in comparison with RealBasicVSR~\cite{chan2022investigating}.
To summarise, our methods have better performance and visual results on RealVSR dataset.
Please refer to supplementary material for more qualitative results.

\subsection{Experimental Results on MVSR4$\times$}
Real-world $\times$4 data has a much more complex degradation than $\times2$ data. 
As shown in Table.~\ref{tab:result}, on MVSR4$\times$, our method EAVSR+ still has state-of-the-art performance between all algorithms on our MVSR4$\times$ dataset. 
And the gap between BasicVSR and EAVSR is increased than that in RealVSR \cite{yang2021real} dataset.
It implies that an optical flow network pre-trained on synthetic data may have performance dropping in real-world data due to the difference in degradation. 
In addition, EDVR \cite{wang2019edvr} has a better performance than BasicVSR on $\times$4 real-world dataset, which is the opposite on the $\times$2 dataset. 
It also demonstrates that when degradation is more severe, the deformable convolution will be more flexible than the optical flow network, and it is not comprehensive to verify the performance of different VSR methods only on the $\times$2 real-world dataset.

Meanwhile, we show the qualitative comparison in Fig. \ref{fig:mvsr4x}.
It can be seen that the seam of other methods between tiles is severely blurry except for EAVSR+ and EAVSR. 
Our EAVSR+ can recover more edges. 
And in Fig~\ref{fig:gan} (b), our EAVSRGAN+ can generate the result more clearly.
%-------------------------------------------------------------------------

%------------------------------------------------------------------------
\section{Ablation Study}
\begin{figure}[t]
% \vspace{-1mm}
    \centering % , height=1.03\linewidth
    \includegraphics[width=\linewidth]{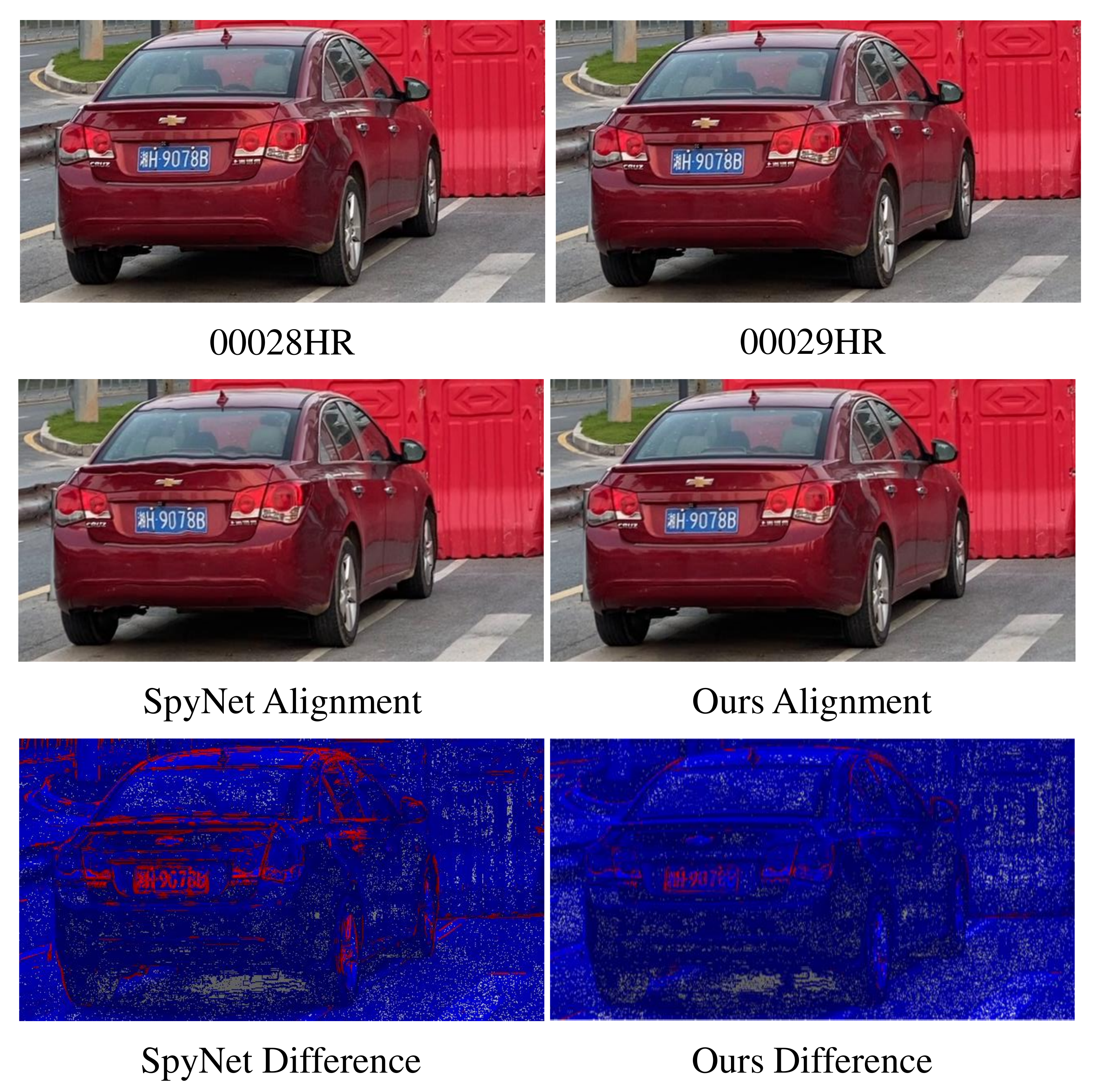}
    \caption{Alignment effect comparison on SPyNet~\cite{ranjan2017optical} and our ResflowNet. Our method can generate a more reliable offset with less distortion.}
    \label{fig:align-result} % is more photo-realistic.
    % \vspace{-10mm}
\end{figure}
In this section, we conduct extensive ablation studies on the proposed MultiAdaSTN and assess the effect of our training loss. 
All experiments are conducted on the RealVSR dataset \cite{yang2021real} with EAVSR+.
\label{sec:abs}

\subsection{Combination of MultiAdaSTN}
\begin{table}[t]
  \small
  \caption{Quantitative comparison with the different alignment combinations in MultiAdaSTN.}
% \vspace{-0.5em}
    \begin{tabular}{cccc}
          \toprule
          ResflowNet&DeformNet& PSNR & SSIM\\
            \midrule
            $\times$&$\times$   & 24.13 & 0.7854\\   
            \checkmark&$\times$  & 24.24 & 0.7902\\
            $\times$&\checkmark     & 24.20 & 0.7876\\
            \checkmark&\checkmark      & 24.41 & 0.7953\\
          \bottomrule
        \end{tabular}
    \centering
    \label{tab:layer}
\end{table}
We verify the performance of ResflowNet and DeformNet in MultiAdaSTN and will give a discussion about the combination of MutiAdaSTN. 
When neither ResflowNet nor DeformNet is applied, we directly leverage the basic offset calculated by pre-trained SPyNet~\cite{ranjan2017optical} to warp the neighboring frames, which is considered the baseline. 
As shown in Table~\ref{tab:layer}, benefiting from the estimation of residual offset by ResflowNet, we get 0.11 dB PSNR gain compared to the baseline. 
Coupled with DeformNet, the PSNR gain of 0.17 dB can be further obtained.
Therefore, we concatenate these two modules together for finer results, and meanwhile will not introduce a large amount of time consumption.

In addition, We also provide the results of SPyNet and our ResflowNet on real-world data alignment in Fig.~\ref{fig:align-result}. 
The alignment result by SPyNet is warped from HR image `00028HR' by calculating the optical flow from LR images  `00028LR' to `00029LR'.  
For ours, the basic offset is computed by SPyNet from `00028LR' to `00029LR' while the residual offset is computed by ResflowNet. 
The alignment result is warped from `00028HR' by combination of basic offset and residual offset. 
`Difference'  means to calculate the difference between the aligned `00028HR' image and `00029HR' image where red areas mean a large difference and blue areas mean a slight difference. 
Obviously, our method aligns better with introducing less distortion.

\subsection{Effect of Training Loss}
\begin{table}[t]
  \small
  \caption{Quantitative comparison with the different data post-processing during training.}
% \vspace{-0.5em}
    \begin{tabular}{cccc}
          \toprule
          \tabincell{c}{Position\\Alignment} & \tabincell{c}{Color\\Correction}
          & PSNR & SSIM\\
            \midrule
            $\times$&$\times$   & 24.32 & 0.7928\\   
            \checkmark&$\times$  & 24.39 & 0.7935\\
            $\times$&\checkmark     & 24.34 & 0.7920\\
            \checkmark&\checkmark      & 24.41 & 0.7953\\
          \bottomrule
        \end{tabular}
    \centering
    \label{tab:loss}
\end{table}
The data pre-processing cannot completely solve the spatial position misalignment and color inconsistency issues.
We leverage a robust optical flow network PWC-Net~\cite{sun2018pwc} to further warp the HR patch and the guided filtering~\cite{he2012guided} to convert the color of HR in the training phase.
To demonstrate the advantages of the data post-processing, we verified the effect of each operation separately. 
Table~\ref{tab:loss} shows the results using different operations, each operation significantly improves the performance compared with the baseline.
Due to space limitations, some qualitative results will show in the supplementary material.

\section{Conclusion}
The lack of real-world paired datasets with large scale factors and the inapplicability of alignment methods set up obstacles for existing video super-resolution (VSR) methods to apply in real scenarios.
For these two issues, we build a $\times$4 real-world VSR dataset (named MVSR4$\times$) and propose an effective alignment method EAVSR.
The low- and high-resolution videos in the MVSR4$\times$ dataset are captured with different focal length lenses of a smartphone, respectively. 
EAVSR takes the proposed multi-layer adaptive spatial transform network (MultiAdaSTN) to refine the offsets provided by the optical flow estimation network.
Extensive experiments show the effectiveness and practicality of our method, and we achieve state-of-the-art performance in the real-world VSR task. 

% \clearpage

%%%%%%%%% REFERENCES
{\small
\bibliographystyle{ieee_fullname}
\bibliography{egbib}
}
\clearpage

\appendix

\setcounter{section}{0}
\setcounter{table}{0}
\setcounter{figure}{0}
\renewcommand\thesection{\Alph{section}}
\renewcommand\thesubsection{\thesection.\arabic{subsection}}
\renewcommand\thefigure{\Alph{figure}}
\renewcommand\thetable{\Alph{table}}
\renewcommand\thetable{\Alph{table}}
%%%%%%%%% BODY TEXT

\twocolumn[
\begin{@twocolumnfalse}
\section*{\centering{\Large Supplementary Material\\[30pt]}}
\end{@twocolumnfalse}
]

\section{Content}
The content of this supplementary material involves:
\begin{itemize}
    \item Network structures of encoder, reconstructor and upsampler in Sec.~\ref{sec:construct}.
    \item Effect of pre-trained optical flow network in Sec.~\ref{sec:pre-trained_flow}.
    \item Effect of training loss in Sec.~\ref{sec:loss}.
    \item More visual comparisons  in Sec.~\ref{sec:visual}.
\end{itemize}

\section{Network Structures of Encoder, Reconstructor and Upsampler}\label{sec:construct}

\begin{table}[t]
  \caption{Structure configuration of the encoder. The kernel size of `Conv' layers is 3$\times$3.}
  \centering\noindent
  \centering%
%   \vspace{-5mm}
  \begin{center}
    \begin{tabular}{c| l}
      \toprule
      \scriptsize\textbf{\#} & \textbf{Layer name(s)}\\
      \hline\hline
        0   & Conv $\left(3,64\right)$, ReLU\\
        \hline
        1   & Conv $\left(64,64\right)$, ReLU\\
        \hline
        2   & Conv $\left(64,128\right)$, ReLU\\
        \hline
        3   & Conv $\left(128,128\right)$, ReLU\\
        \hline
        4   & Conv $\left(128,256\right)$, ReLU\\
        \hline
      \bottomrule
    \end{tabular}
    \end{center}
  \label{tab:encoder}
\end{table}

\begin{table}[t!]
  \caption{Structure configuration of the combination of reconstructor and upsampler. The kernel size of `Conv' layers is 3$\times$3 and the kernel size of `Conv1$\times$1' layer is 1$\times$1. The negative slope of LeakyReLU is 0.1.}
  \centering\noindent
  \centering%
%   \vspace{-5mm}
  \begin{center}
    \begin{tabular}{c| l}
      \toprule
      \scriptsize\textbf{\#} & \textbf{Layer name(s)}\\
      \hline\hline
        0   & Concat $\left[\rm{LR}, \rm{Propagated\  Feature}\right]$\\
        \hline
        1   & Conv $\left(67,64\right)$, LeakyReLU\\
        \hline
        2   & RCAB $\times$ 30\\
        \hline
        3   & Conv 1$\times$1 $\left(64,64\right)$, LeakyReLU\\
        \hline
        4   & Conv $\left(64,256\right)$, PixelShuffle, LeakyReLU\\
        \hline
        5   & Conv $\left(64,64\right)$, LeakyReLU\\
        \hline
        6   & Conv $\left(64,3\right)$\\
        \hline
        7   & BilinearUpsample $\left(\rm{LR},2\right)$\\
        \hline
        8   & ElementwiseAdd $\left(\#6,\#7\right)$\\
        \hline
      \bottomrule
    \end{tabular}
    \end{center}
  \label{tab:reconstruct}
\end{table}

To estimate the effectiveness of the proposed method comprehensively, we select bidirectional~\cite{chan2021basicvsr} and second-order grid~\cite{chan2021basicvsr++} as our temporal propagation scheme, named EAVSR and EAVSR+, respectively.
Table~\ref{tab:encoder} shows the detailed architectures of the encoder in EAVSR and EAVSR+.
Table~\ref{tab:reconstruct} shows the detailed architectures of  the reconstructor and upsampler in EAVSR.
`Propagated Feature' in Table~\ref{tab:reconstruct} refers to the well-aligned features propagated from neighboring frames, and RCAB denotes the residual channel attention block~\cite{zhang2018image}.
The reconstructor and upsampler architectures of EAVSR+ follow  BasicVSR++~\cite{chan2021basicvsr++}.

\section{Effect of Pre-Trained Optical Flow Network} \label{sec:pre-trained_flow}

\begin{table}[t]
  \small
  \caption{Quantitative comparison with the different pre-trained optical flow networks on RealVSR~\cite{yang2021real} dataset.}
% \vspace{-0.5em}
    \begin{tabular}{ccccc}
          \toprule
          \tabincell{c}{Optical Flow\\Network} & ResflowNet & DeformNet & PSNR & SSIM \\
            \midrule
            SPyNet~\cite{ranjan2017optical}&$\times$&$\times$   & 24.13 & 0.7854\\ 
            SPyNet~\cite{ranjan2017optical}&\checkmark&$\times$  & 24.24 & 0.7902\\
            SPyNet~\cite{ranjan2017optical}&\checkmark&\checkmark      & 24.41 & 0.7953\\
            PWC-Net~\cite{sun2018pwc}&$\times$&$\times$   & 24.19 & 0.7847\\ 
            PWC-Net~\cite{sun2018pwc}&\checkmark&$\times$  & 24.26 & 0.7906\\
            PWC-Net~\cite{sun2018pwc}&\checkmark&\checkmark      & 24.44 & 0.7939\\
          \bottomrule
        \end{tabular}
        % \vspace{2mm}
    \centering
    \label{tab:optical}
\end{table}

\begin{figure*}[t]
    \centering
    \includegraphics[width=\linewidth]{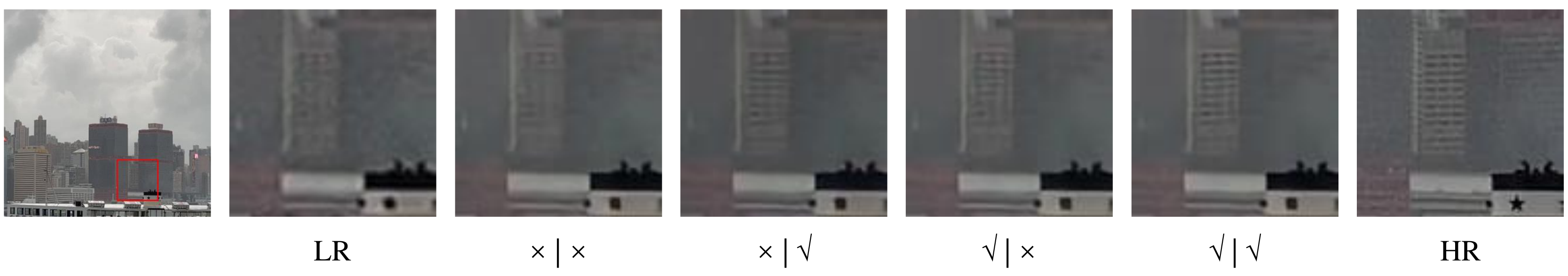}
    \vspace{-6mm}
    \caption{Visual comparison between different data post-processing strategies during training. The first $\checkmark$ or × means using color correction or not, and the second $\checkmark$ or × means using spatial position alignment or not. Please zoom in for details.}
    \label{fig:loss}
    % \vspace{-2mm}
\end{figure*}

\begin{figure*}[t]
    \centering
    \includegraphics[width=\linewidth]{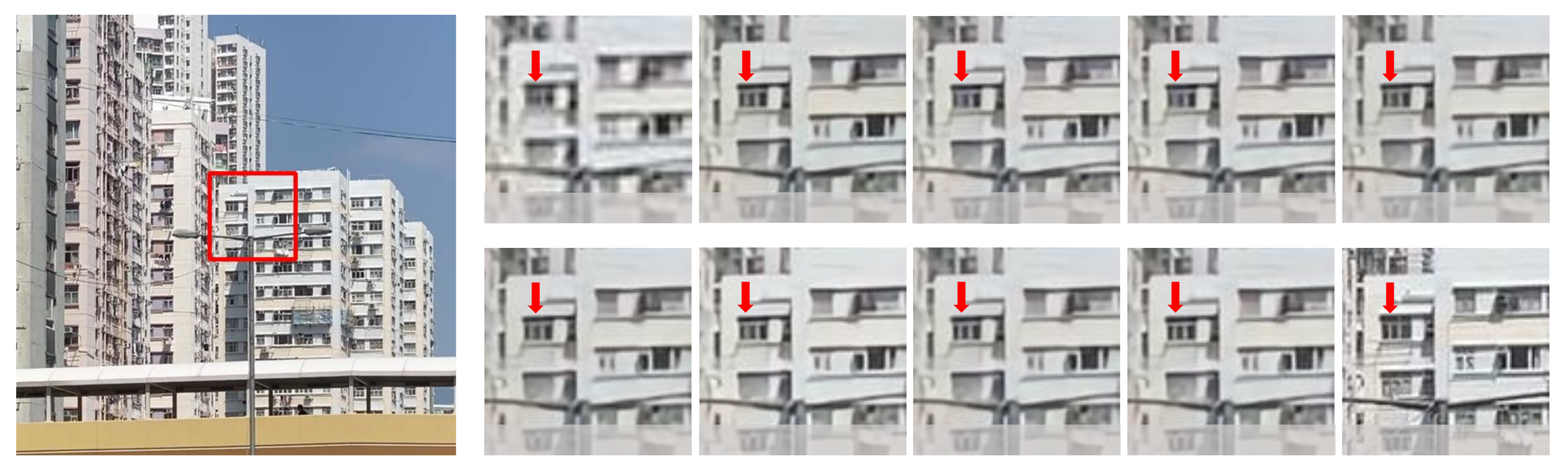}
    \put(-316, 79){\small{LR}}
    \put(-263, 79.5){\small{EDVR~\cite{wang2019edvr}}}
    \put(-196, 79.5){\small{MANA~\cite{MANA}}}
    \put(-132, 79.5){\small{BasicVSR~\cite{chan2021basicvsr}}}
    \put(-60, 79.5){\small{TTVSR~\cite{Liu_2022_CVPR}}}
    \put(-330, 6){\small{ETDM~\cite{Isobe_2022_CVPR}}}
    \put(-272.5, 6){\small{BasicVSR++~\cite{chan2021basicvsr++}}}
    \put(-189, 6){\small{EAVSR}}
    \put(-124, 6){\small{EAVSR+}}
    \put(-45, 6){\small{HR}}
    % \vspace{-2mm}
    \caption{Visual comparison on RealVSR dataset~\cite{yang2021real}. Our methods EAVSR+ and EAVSR can better recover the window contours, especially the window in the upper left corner. Please zoom in for details.}
    \label{fig:realvsr-result1}
    % \vspace{-2mm}
\end{figure*}
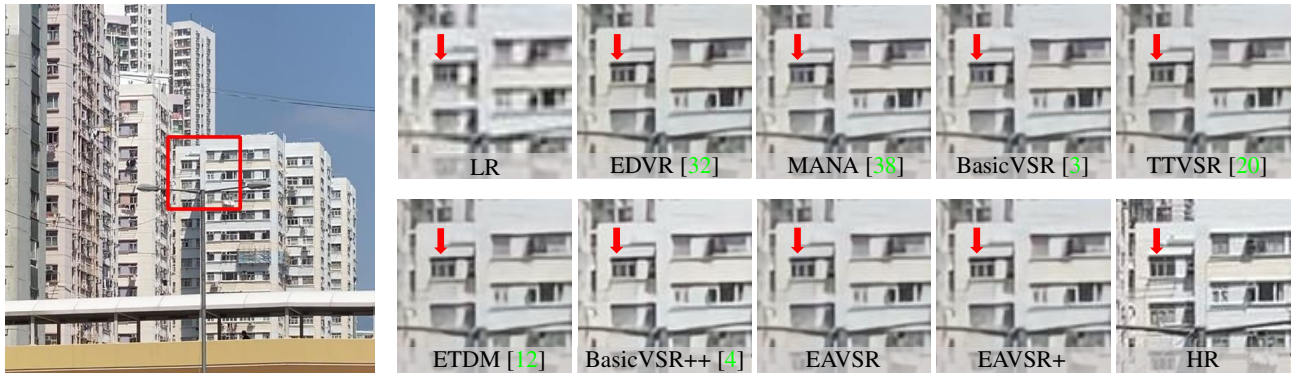

As described in Sec. \textcolor{red}{4.2} of the main text, we follow BasicVSR~\cite{chan2021basicvsr} and BasicVSR++~\cite{chan2021basicvsr++}, using a light-weight pre-trained optical flow network SPyNet~\cite{ranjan2017optical} to calculate the basic offset between neighboring frames.
In this section, we replace SPyNet with PWC-Net~\cite{sun2018pwc} to verify the effect of the proposed ResflowNet and DeformNet when taking a better and more robust pre-trained optical flow network. 

Table~\ref{tab:optical} shows the experiment results on  RealVSR dataset~\cite{yang2021real}. 
The first and the fourth rows demonstrate that despite deploying a better optical flow network, the final performance still has a limited improvement on real-world data. 
Taking our ResflowNet to compensate for the error caused by the pre-trained flow network, the second and fifth rows show that the performance promotes significantly.
Moreover, the third and the last lines indicate the validity of our DeformNet as well.

\section{Effect of Training Loss} \label{sec:loss}

In this section, we show some visual results in Fig~\ref{fig:loss} when using different data post-processing strategies during training, which is mentioned in Sec. \textcolor{red}{6.2} of the main text.
When neither color correction nor spatial position alignment is applied, the results cannot restore the correct details.
When we use the PWC-Net~\cite{sun2018pwc} to mitigate the spatial misalignment between LR and HR, the result can generate more details.
Moreover, if we only apply the guided image filtering~\cite{he2012guided}  to correct the color of HR, the result can keep the brightness consistent with LR but lead to blurry. 
Utilizing both color correction and spatial position alignment  can recover more textures and remain color consistent with the LR image.

\section{Visual Comparison} \label{sec:visual}
\begin{figure*}[t]
    \centering
    \includegraphics[width=\linewidth]{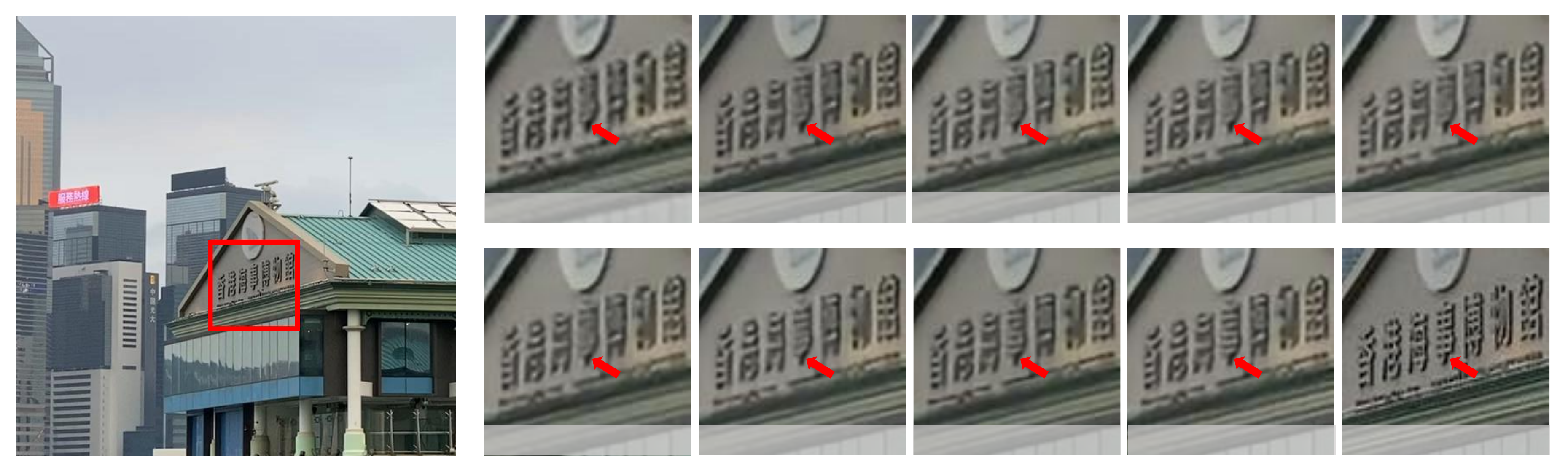}
    \put(-316, 79){\small{LR}}
    \put(-263, 79.5){\small{EDVR~\cite{wang2019edvr}}}
    \put(-196, 79.5){\small{MANA~\cite{MANA}}}
    \put(-132, 79.5){\small{BasicVSR~\cite{chan2021basicvsr}}}
    \put(-60, 79.5){\small{TTVSR~\cite{Liu_2022_CVPR}}}
    \put(-330, 6){\small{ETDM~\cite{Isobe_2022_CVPR}}}
    \put(-272.5, 6){\small{BasicVSR++~\cite{chan2021basicvsr++}}}
    \put(-189, 6){\small{EAVSR}}
    \put(-124, 6){\small{EAVSR+}}
    \put(-45, 6){\small{HR}}
    \caption{Visual comparison on RealVSR dataset~\cite{yang2021real}. Our methods EAVSR+ and EAVSR can  recover the characters better. Please zoom in for details.}
    \label{fig:realvsr-result2}
    % \vspace{-4mm}
\end{figure*}

\begin{figure*}[t]
    \centering
    \includegraphics[width=\linewidth]{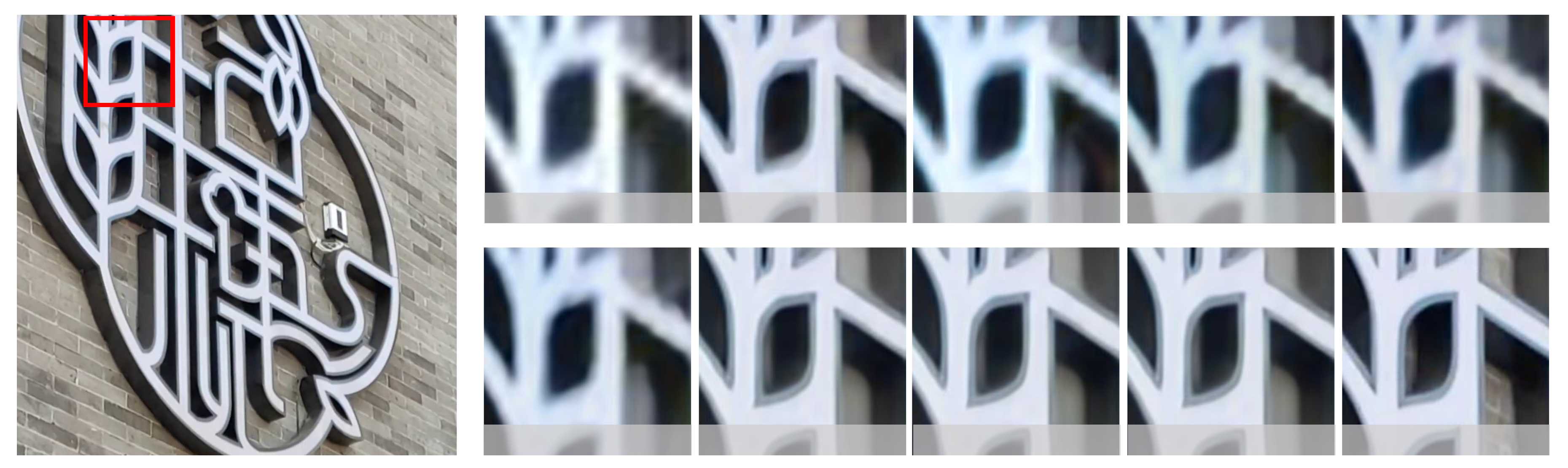}
    \put(-316, 79){\small{LR}}
    \put(-263, 79.5){\small{EDVR~\cite{wang2019edvr}}}
    \put(-196, 79.5){\small{MANA~\cite{MANA}}}
    \put(-132, 79.5){\small{BasicVSR~\cite{chan2021basicvsr}}}
    \put(-60, 79.5){\small{TTVSR~\cite{Liu_2022_CVPR}}}
    \put(-330, 6){\small{ETDM~\cite{Isobe_2022_CVPR}}}
    \put(-272.5, 6){\small{BasicVSR++~\cite{chan2021basicvsr++}}}
    \put(-189, 6){\small{EAVSR}}
    \put(-124, 6){\small{EAVSR+}}
    \put(-45, 6){\small{HR}}
    \caption{Visual comparison on MVSR4$\times$ dataset. Our methods EAVSR+ and EAVSR can generate clearer contours with fewer artifacts. Please zoom in for details.}
    \label{fig:mvsr4x-result1}
    % \vspace{-4mm}
\end{figure*}

\begin{figure*}[t]
    \centering
    \includegraphics[width=\linewidth]{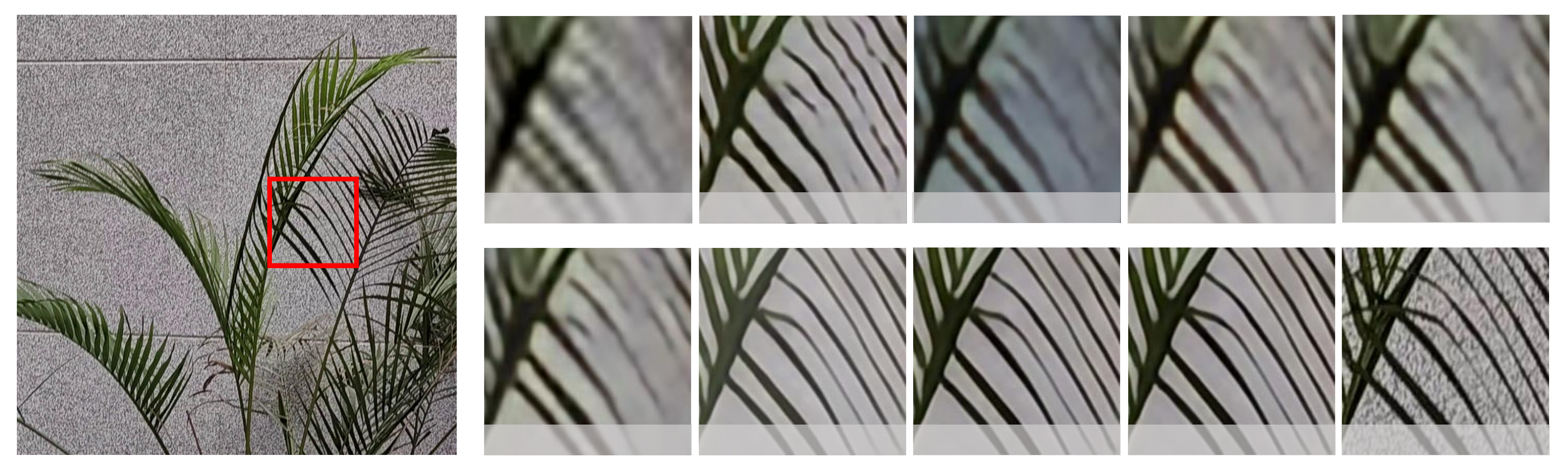}
    \put(-316, 79){\small{LR}}
    \put(-263, 79.5){\small{EDVR~\cite{wang2019edvr}}}
    \put(-196, 79.5){\small{MANA~\cite{MANA}}}
    \put(-132, 79.5){\small{BasicVSR~\cite{chan2021basicvsr}}}
    \put(-60, 79.5){\small{TTVSR~\cite{Liu_2022_CVPR}}}
    \put(-330, 6){\small{ETDM~\cite{Isobe_2022_CVPR}}}
    \put(-272.5, 6){\small{BasicVSR++~\cite{chan2021basicvsr++}}}
    \put(-189, 6){\small{EAVSR}}
    \put(-124, 6){\small{EAVSR+}}
    \put(-45, 6){\small{HR}}
    \caption{Visual comparison on MVSR4$\times$ dataset. Our methods EAVSR and EAVSR+ can restore sharper branches. Please zoom in for details.}
    \label{fig:mvsr4x-result2}
    % \vspace{-4mm}
\end{figure*}

\begin{figure*}[t]
    \centering
    \includegraphics[width=\linewidth]{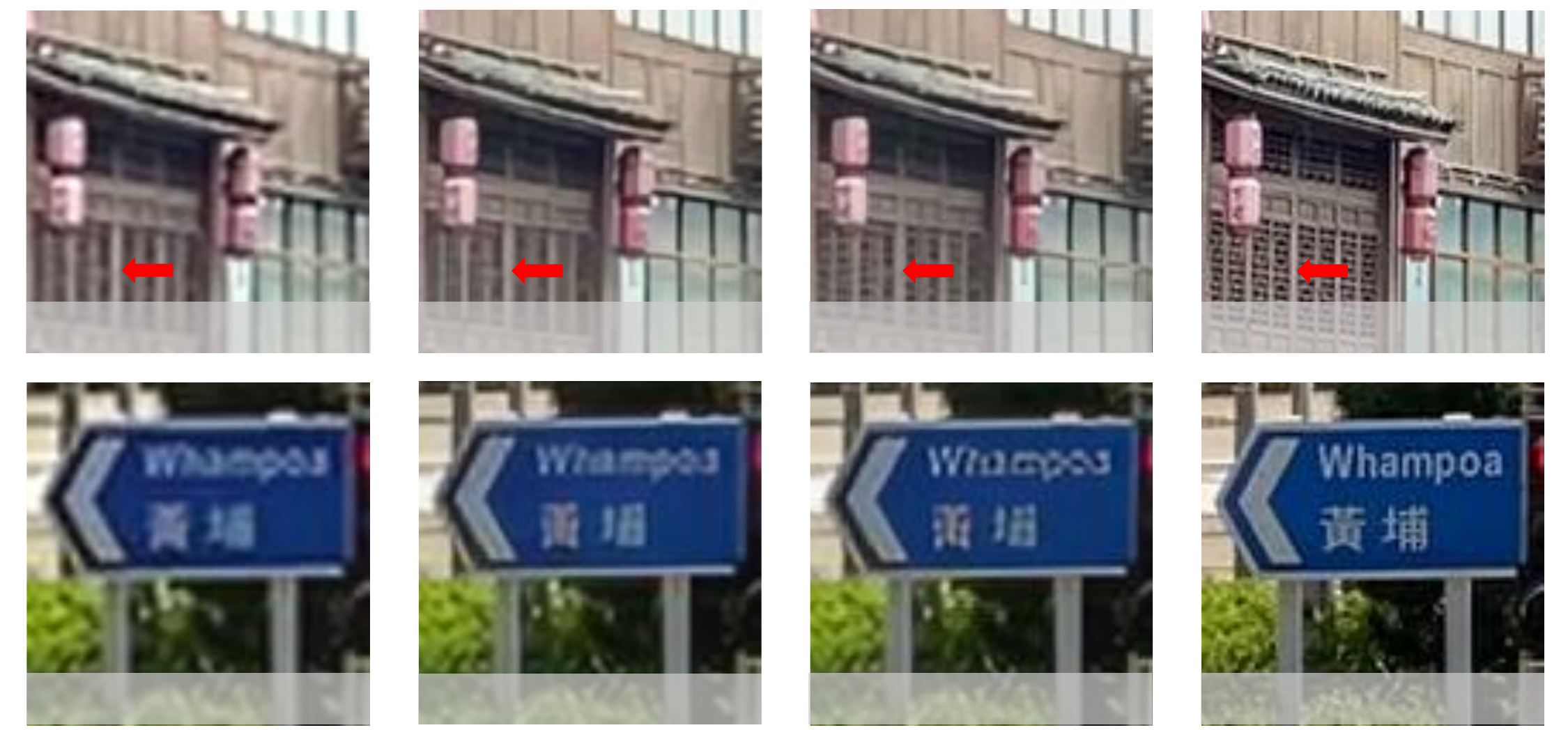}
    \put(-440, 124.5){{LR}}
    \put(-347, 124.5){RealBasicVSR~\cite{chan2022investigating}}
    \put(-214, 124.5){EAVSRGAN+}
    \put(-68.5, 124.5){HR}
    \put(-440, 6){LR}
    \put(-347, 6){RealBasicVSR~\cite{chan2022investigating}}
    \put(-214, 6){EAVSRGAN+}
    \put(-68.5, 6){HR}
    \caption{Visual comparison on RealVSR~\cite{yang2021real} dataset between methods trained with adversarial loss. Our results from EAVSRGAN+ have more details and are more photo-realistic. Please zoom in for more details.}
    \label{fig:realvsrgan-result}
    % \vspace{-4mm}
\end{figure*}

\begin{figure*}[t]
    \centering
    \includegraphics[width=\linewidth]{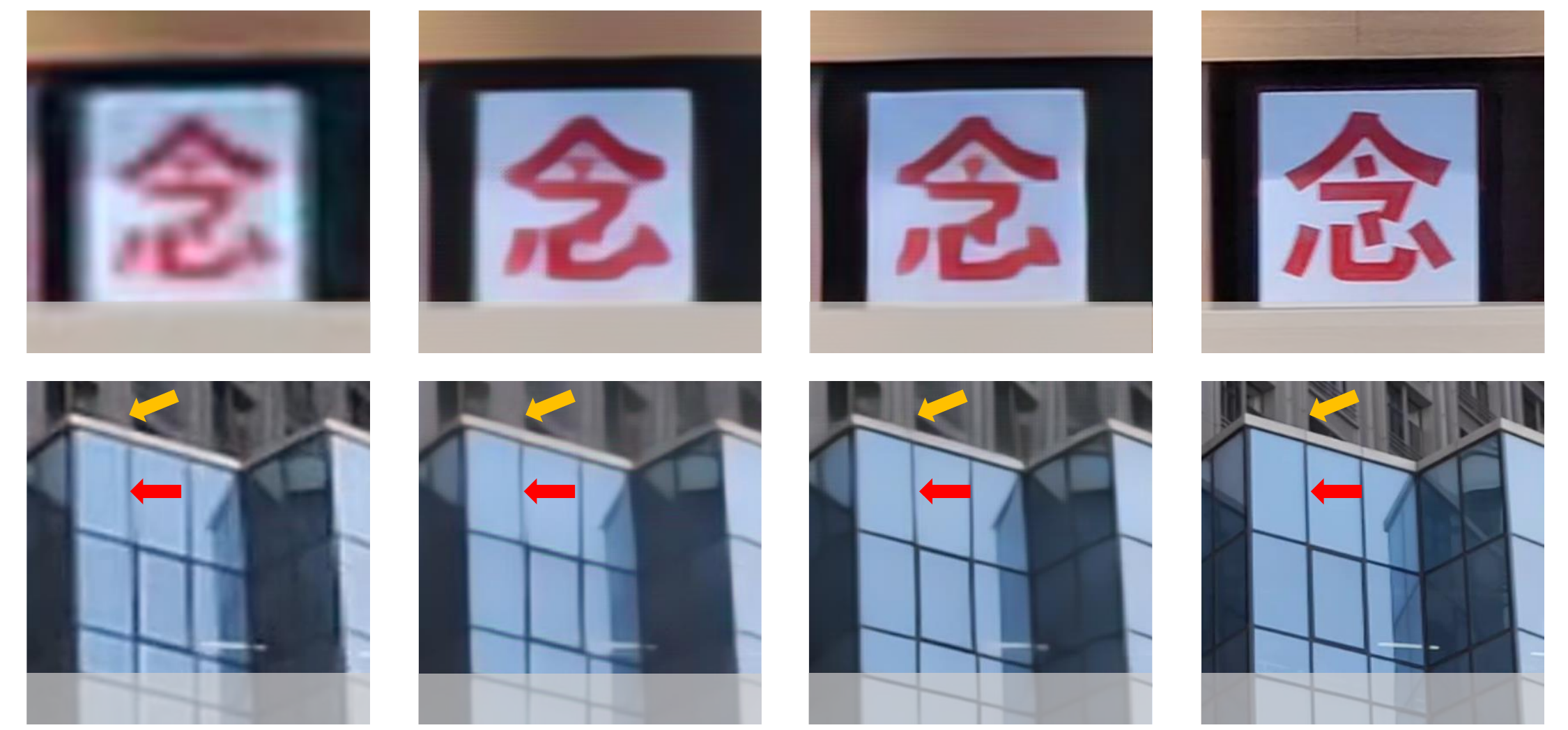}
    \put(-440, 124.5){{LR}}
    \put(-347, 124.5){RealBasicVSR~\cite{chan2022investigating}}
    \put(-214, 124.5){EAVSRGAN+}
    \put(-68.5, 124.5){HR}
    \put(-440, 6){LR}
    \put(-347, 6){RealBasicVSR~\cite{chan2022investigating}}
    \put(-214, 6){EAVSRGAN+}
    \put(-68.5, 6){HR}
    \caption{Visual comparison on MVSR4$\times$ dataset between methods trained with adversarial loss. Our results from EAVSRGAN+ have clearer edges and are more photo-realistic. Please zoom in for more details.}
    \label{fig:mvsr4xgan-result}
    % \vspace{-4mm}
\end{figure*}
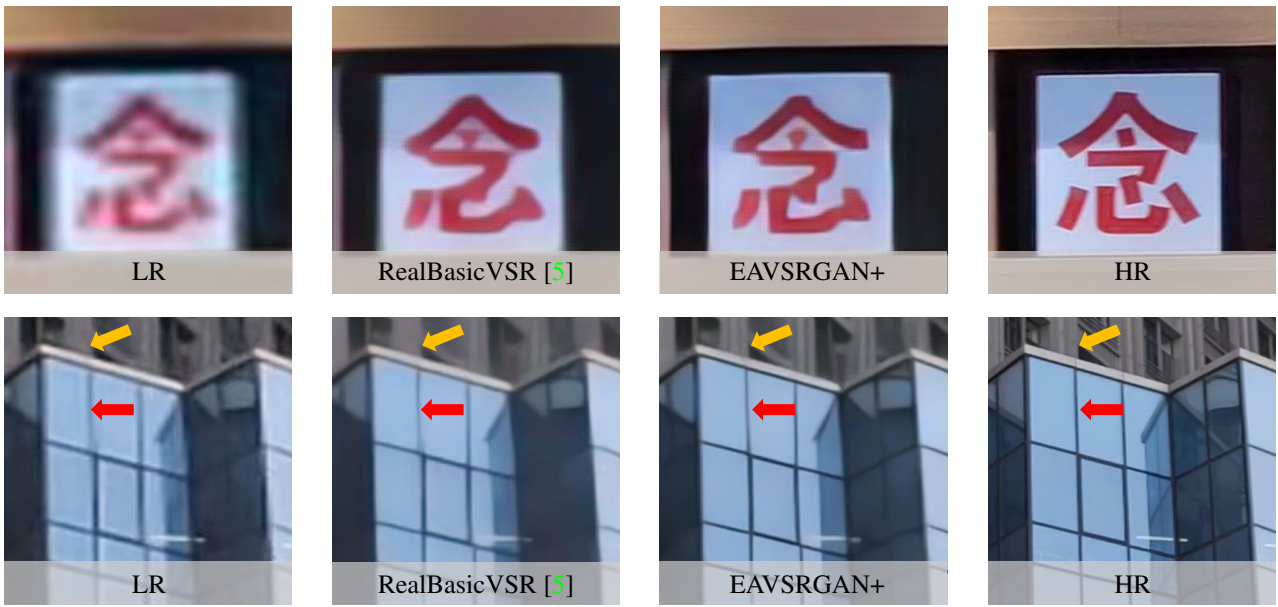

In this section, we provide more qualitative comparisons between our methods and other state-of-the-art algorithms on RealVSR~\cite{yang2021real} dataset (see Fig.~\ref{fig:realvsr-result1}, Fig.~\ref{fig:realvsr-result2} and Fig.~\ref{fig:realvsrgan-result}), and MVSR4$\times$ dataset (see Fig.~\ref{fig:mvsr4x-result1},  Fig.~\ref{fig:mvsr4x-result2} and Fig.~\ref{fig:mvsr4xgan-result}), respectively.
Specifically, we show results of methods trained only with $\ell_1$ loss in Figs~\ref{fig:realvsr-result1}$\sim$\ref{fig:mvsr4x-result2}, and methods trained with additional adversarial loss in Figs~\ref{fig:realvsrgan-result}$\sim$\ref{fig:mvsr4xgan-result}, respectively.

\clearpage

\end{document}